\documentclass{article}

\usepackage{arxiv}

\usepackage[utf8]{inputenc} 
\usepackage[T1]{fontenc}    
\usepackage{lmodern}
\usepackage{hyperref}       
\usepackage{url}            
\usepackage{booktabs}       
\usepackage{amsfonts}       
\usepackage{nicefrac}       
\usepackage{microtype}      
\usepackage{amsmath}
\usepackage{bm}
\usepackage{mathtools}
\usepackage[labelfont=bf]{caption}
\usepackage{multirow}
\usepackage{adjustbox}
\usepackage{natbib}

\usepackage{filecontents}

\usepackage[normalem]{ulem}

\title{Fast Maximum Likelihood Estimation and Supervised\protect\\Classification for the Beta-Liouville Multinomial}

\author{
  Steven M.~Lakin \\
  Department of Microbiology, Immunology and Pathology\\
  College of Veterinary Medicine and Biomedical Sciences \\
  Colorado State University\\
  Fort Collins, CO 80523 \\
  \texttt{steven.lakin@colostate.edu} \\
  \\
  \textbf{Zaid Abdo} \\
  Department of Microbiology, Immunology and Pathology\\
  College of Veterinary Medicine and Biomedical Sciences \\
  Colorado State University\\
  Fort Collins, CO 80523 \\
  \texttt{zaid.abdo@colostate.edu} \\}

\date{}

\begin{document}
\maketitle

\begin{abstract}
The multinomial and related distributions have long been used to model categorical, count-based data in fields ranging from bioinformatics to natural language processing. Commonly utilized variants include the standard multinomial and the Dirichlet multinomial distributions due to their computational efficiency and straightforward parameter estimation process. However, these distributions make strict assumptions about the mean, variance, and covariance between the categorical features being modeled.  If these assumptions are not met by the data, it may result in poor parameter estimates and loss in accuracy for downstream applications like classification.  Here, we explore efficient parameter estimation and supervised classification methods using an alternative distribution, called the Beta-Liouville multinomial, which relaxes some of the multinomial assumptions.  We show that the Beta-Liouville multinomial is comparable in efficiency to the Dirichlet multinomial for Newton-Raphson maximum likelihood estimation, and that its performance on simulated data matches or exceeds that of the multinomial and Dirichlet multinomial distributions.  Finally, we demonstrate that the Beta-Liouville multinomial outperforms the multinomial and Dirichlet multinomial on two out of four gold standard datasets, supporting its use in modeling data with low to medium class overlap in a supervised classification context.
\end{abstract}

\section{Introduction}
Count data arise in many contexts, including in natural language processing, ecology, and bioinformatics (\citealt{wu2017,valpine2013,lewis1998}).  For example, data such as the words in an email could be features that are counted and used to classify each email into a category (e.g. spam or ham).  While many probability distributions can be used to model categorical count data, the multinomial and Dirichlet multinomial (DM) distributions have been shown to perform well and are widely used for this task (\citealt{bouguila2008}).  The simplicity of these distributions enables their integration with computationally efficient classifiers like naïve Bayes, making them popular for large-scale classification tasks. For example, one can obtain estimates of multinomial parameters over any number of features by simple division, circumventing the need for more complicated parameter estimation techniques.  Additionally, the Dirichlet distribution is a convenient prior distribution for the multinomial, resulting in "smoother" multinomial estimates that help to prevent overfitting of the training data (\citealt{kaur2014}). \par

However, the multinomial and DM distributions assume strict negative covariance between features, making them ill-suited for modeling data that violate this assumption.  Since features often co-occur in practice, these distributions may not be optimal for capturing nuanced feature relationships in categorical datasets, such as semantics in text processing (\citealt{lilleberg2015}). These distributions also have means and variances for each feature that are linked due to sharing the same parameters, much like the Poisson distribution.  As a result, features with truly different variances but the same mean will be modeled identically, which can further confuse classifiers. To address these issues with the multinomial and DM distributions without sacrificing computational efficiency, the Beta-Liouville distribution can be used as a conjugate prior to the multinomial distribution. Utilizing a Beta-Liouville prior gives rise to the Beta-Liouville multinomial (BLM) distribution (\citealt{gupta2001}). The BLM relaxes the previously mentioned assumptions by introducing an additional parameter over the DM distribution, which may improve modeling results. \par 

Recent applications of the BLM distribution have been explored by Bouguila (\citeyear{bouguila2008,bouguila2010,bouguila2013}), particularly in computer vision and natural language processing.  Bouguila’s work has focused on unsupervised clustering and semi-supervised classification using methods such as expectation maximization, variational learning with finite mixture models, and online learning.  Bouguila’s group has shown that the BLM distribution improves upon the DM distribution for certain datasets where class labels are determined by the learning process, such as in clustering via expectation maximization (\citealt{bouguila2010}).  However, their work has not applied the BLM distribution to strictly supervised problems, in which the true class labels are fixed and not determined by the learning process.  Additionally, their work does not thoroughly explore the computational efficiency and stability of learning BLM-based models from observed data.  In this manuscript, we extend Bouguila’s work by applying the BLM distribution to supervised classification, focusing on efficient computation, stability of parameter optimization, and classification power. \par 

Previous work by Sklar (\citeyear{sklar2014}) provided an efficient implementation of Newton-Raphson maximum likelihood estimation (MLE) for the DM distribution. Building on the framework introduced by Sklar, we present a fast MLE implementation for the BLM and show the necessary conditions for its convergence during Newton-Raphson optimization.  We then explore conditions under which Sklar’s approach is computationally efficient compared to vectorized and approximate versions of the Newton-Raphson algorithm.  Next, we perform a power analysis for the BLM distribution using simulated data. Finally, we evaluate the accuracy of the DM and BLM against gold standard datasets using a variety of classification techniques.  Overall, we show that both the DM and BLM can be computationally efficient during MLE, and the DM and BLM MLEs outperform the multinomial for datasets with lower levels of class overlap, while the multinomial distribution demonstrates comparable or better accuracy otherwise.

\section{Methods}
\subsection{Definitions}
\label{sec:definitions}

We provide the following definitions to simplify derivations for the Dirichlet Multinomial (DM) and Beta-Liouville Multinomial (BLM) distributions (\citealt{sklar2014}). The gamma function of a strictly positive integer $a$ is,

\begin{align*}
    \Gamma (a) = (a - 1)!
\end{align*}

The dual-input gamma function for strictly positive integers $a$ and $b$ is,

\begin{align*}
    \Gamma (a, b) = \frac{\Gamma (a + b)}{\Gamma (a)} = {\displaystyle \prod_{i =0}^{b-1} (a + i)}
\end{align*}

The dual-input log-gamma function is the log of the dual-input gamma function,

\begin{align*}
    \ln \Gamma (a, b) = {\displaystyle \sum_{i =0}^{b-1} \ln(a + i)}
\end{align*}

We also define $\underset{N \times (D+1)}{\bm{X}}$ to be an observed data matrix of non-negative integers. $\bm{X}$ contains $N$ observations of a $D+1$ dimensional data vector $\bm{x}$, where $D+1$ is the number of categories being counted (Table 1). 

\begin{table}[!htbp]
 \caption{Example of the data matrix $\bm{X}$ with $N$ observations and categories $d = 1 \dots D+1$}
  \centering
  \begin{tabular}{c|cccc}
    & \multicolumn{4}{c}{Category} \\
    \midrule
    Observation & $d = 1$ & $d = 2$ & \dots & $D+1$ \\
    \midrule
    $n = 1$ & 3 & 17 & \dots & 8     \\
    $n = 2$ & 0 & 23 & \dots & 11     \\
    \vdots & \vdots & \vdots & $\ddots$ & \vdots     \\
    $N$ & 20 & 0 & \dots & 9 \\
  \end{tabular}
  \label{tab:table}
\end{table}

\subsection{Derivations for the Beta-Liouville Multinomial distribution and the associated likelihood function}
Liouville distributions of the second kind have $D$ parameters ($\alpha_1, \alpha_2, \dots, \alpha_{D}$) and additional parameters $\xi$ associated with a generating density function $f(\cdot)$ (\citealt{bouguila2010}),

\begin{equation}\label{eq1}
    P(\bm{p}|\alpha_1, \dots, \alpha_{D}, \xi) = f(u|\xi) \frac{\Gamma(\sum_{d=1}^D \alpha_d)}{u^{\sum_{d=1}^D \alpha_d-1}} {\displaystyle \prod_{d = 1}^D \frac{p_d^{\alpha_d-1}}{\Gamma(\alpha_d)}}
\end{equation}

where $u = \sum_{d=1}^D p_d < 1, p_d > 0, d=1\dots D$, and $D$ as defined above. The Beta-Liouville distribution is formed when using the beta distribution with parameters $\xi = \{\alpha, \beta\}$ as the generating density function for $u$, resulting in,

\begin{equation}\label{eq2}
    P(\bm{p}|\bm{\theta}) = \frac{\Gamma(\sum_{d=1}^D \alpha_d)\Gamma(\alpha + \beta)}{\Gamma(\alpha) \Gamma(\beta)} u^{\alpha - \sum_{d=1}^D \alpha_d} (1-u)^{\beta-1} {\displaystyle \prod_{d=1}^D \frac{p_d^{\alpha_d-1}}{\Gamma(\alpha_d)}}
\end{equation}

where $\bm{\theta} = (\alpha_1, \dots, \alpha_D, \alpha, \beta)$. Unlike the Dirichlet distribution, which has $D+1$ parameters, equal to the number of observed categories, the Beta-Liouville distribution has $D+2$ parameters. Similar to the Dirichlet, the Beta-Liouville distribution presented in equation \ref{eq2} can be used as a conjugate prior to the multinomial distribution resulting in the BLM distribution,

\begin{equation}\label{eq3}
    P(\bm{x}, \bm{p}|\bm{\theta}) = \left[\frac{\Gamma(\sum_{d=1}^{D} \alpha_d) \Gamma(\alpha+\beta)}{\Gamma(\alpha)\Gamma(\beta)\prod_{d=1}^{D}\Gamma(\alpha_d)}\left({\displaystyle \sum_{d=1}^{D} p_d}\right)^{\alpha' - \sum_{d=1}^{D} \alpha_d'}\right] \left[\frac{\Gamma((\sum_{d=1}^{D+1} x_d)+1)}{\prod_{d=1}^{D+1} \Gamma(x_d+1)} \left(1 - {\displaystyle \sum_{d=1}^{D} p_d } \right)^{\beta' -1} {\displaystyle \prod_{d=1}^{D} p_d^{\alpha_d' -1}} \right]
\end{equation}

where $\alpha_d' = \alpha_d + x_d, \alpha' = \alpha + \sum_{d=1}^{D} x_d$, and $\beta' = \beta + x_{D+1}$, following Bouguila's notation (Bouguila 2011). Marginalizing over $\bm{p}$ and assuming that the data vector $\bm{x}$ has been observed results in the likelihood function,

\begin{align}\label{eq4}
    P(\bm{x}|\bm{\theta}) & = \int_{\bm{p}}  P(\bm{x}, \bm{p}|\bm{\theta)} d\bm{p} \nonumber\\ 
    & = \frac{\Gamma\left(1 + \sum_{d=1}^{D+1} x_d\right) \Gamma\left(\sum_{d=1}^{D}\alpha_d\right) \Gamma(\alpha+\beta) \Gamma\left(\alpha+\sum_{d=1}^{D} x_d\right) \Gamma(\beta+x_{D+1}){\displaystyle \prod_{d=1}^{D} \Gamma(\alpha_d+x_d)}}{{\displaystyle \left(\prod_{d=1}^{D+1} \Gamma(x_d+1)\right)} \Gamma\left(\sum_{d=1}^{D} \left(\alpha_d + x_d\right)\right) \Gamma\left(\alpha+\beta+\sum_{d=1}^{D+1} x_d\right) \Gamma(\alpha) \Gamma(\beta) {\displaystyle \prod_{d=1}^{D} \Gamma(\alpha_d)}}
\end{align}

Utilizing \ref{eq4} and the dual-input gamma function, introduced above, the log-likelihood function can be written as follows for the entire data matrix $\bm{X}$,

\begin{align}\label{eq5}
    \ell(\bm{\theta}) & = \ln(P(\bm{X}|\bm{\theta})) \nonumber\\
    \begin{split}
         = {\displaystyle \sum_{n=1}^{N} \left[ \ln\Gamma(1, \sum_{d=1}^{D+1} x_{nd}) + \ln\Gamma\left( \alpha, \sum_{d=1}^D x_{nd} \right) + \ln\Gamma(\beta, x_{n,D+1}) - \ln\Gamma\left(\sum_{d=1}^{D} \alpha_d, \sum_{d=1}^{D} x_{nd}\right) \right]} \\
        {\displaystyle + \sum_{n=1}^{N} \left[  \left(\sum_{d=1}^{D} \ln\Gamma(\alpha_d, x_{nd})\right) - \ln\Gamma(\alpha + \beta,\sum_{d=1}^{D+1} x_{nd}) - \left(\sum_{d=1}^{D+1} \ln\Gamma(1, x_{nd})\right) \right]}
        \end{split}
\end{align}

Utilizing the definition of the dual-input log-gamma function and rearranging equation \ref{eq5}, removing constant terms that do not involve the parameters, we have,

\begin{align}\label{eq6}
        \ell(\bm{\theta}) & \sim {\displaystyle \sum_{n=1}^N \left[ \sum_{d=1}^{D} \sum_{i=0}^{x_{nd}-1} \ln(\alpha_d + i) - \sum_{i=0}^{\left(\sum_D x_{nd}\right) -1} \ln\left(\sum_{d=1}^{D} \alpha_d + i\right) \right] } \nonumber\\
       & {\displaystyle + \sum_{n=1}^N \left[ \sum_{i=0}^{(\sum_D x_{nd})-1} \ln( \alpha + i) + \sum_{i=0}^{\left(x_{n,D+1}\right)-1} \ln(\beta + i) - \sum_{i=0}^{\left(\sum_{D+1} x_{nd}\right)-1} \ln(\alpha + \beta + i)\right] }
\end{align}

We seek to find the maximum likelihood estimates (MLEs) of $\bm{\theta}$ that optimize \ref{eq6},

\begin{align*}
 MLE(\bm{\theta}) = \underset{\bm{\theta}}{\text{argmax }} \ell(\bm{\theta})
\end{align*}

using the following optimization approaches.

\subsection{MLE of the Beta-Liouville Multinomial distribution}
In what follows, we use the Newton-Raphson algorithm for optimization, with step as in the following equation, utilizing the log-likelihood function \ref{eq6} to find the MLEs.

\begin{equation}\label{eq7}
 \bm{\theta_}{\text{new}} = \bm{\theta}_{\text{old}} - \bm{H}^{-1}\bm{g}
\end{equation}

where $\bm{g}$ is the gradient vector and $\bm{H}$ is the Hessian matrix. The Hessian matrix is block diagonal (\citealt{eves1980}) and its inverse is represented as follows, 

\begin{equation}\label{eq8}
 \bm{H}(\ell(\bm{\theta}))^{-1} = \text{block-diag}\left\{\bm{H}(\ell(\alpha_1, \dots, \alpha_D))^{-1}, \bm{H}(\ell(\alpha, \beta))^{-1} \right\}
\end{equation}

The gradient and Hessian of \ref{eq6} are therefore,

\begin{align*}
 g_{\alpha_d} = {\displaystyle \sum_{n=1}^N \left[ \sum_{i=0}^{x_{nd}-1} (\alpha_d + i)^{-1} - \sum_{i=0}^{(\sum_D x_{nd})-1} \left( \sum_{d=1}^D \alpha_d + i \right)^{-1}  \right], \text{ for } d \in 1 \dots D }
\end{align*}

\begin{equation}\label{eq9}
 g_{\alpha} =  {\displaystyle \sum_{n=1}^N \left[ \sum_{i=0}^{(\sum_D x_{nd})-1} (\alpha + i)^{-1} - \sum_{i=0}^{(\sum_{D+1} x_{nd})-1} (\alpha + \beta + i)^{-1} \right] }
\end{equation}

\begin{align*}
 g_{\beta} = {\displaystyle \sum_{n=1}^N \left[ \sum_{i=0}^{x_{n(D+1)}-1} (\beta + i)^{-1} - \sum_{i=0}^{(\sum_{D+1} x_{nd})-1} (\alpha + \beta + i)^{-1} \right] }
\end{align*}

\begin{equation}\label{eq10}
 \bm{H}(\ell(\alpha_1, \dots, \alpha_D)) =  {\displaystyle \sum_{n=1}^N }
 \begin{bmatrix}
  {\displaystyle c_{\alpha_d} - \sum_{i=0}^{x_{n1}-1} (\alpha_1 + i)^{-2} } & \dots & {\displaystyle c_{\alpha_d} } \\
  \vdots & \ddots & \vdots \\
  {\displaystyle c_{\alpha_d} } & \dots & {\displaystyle c_{\alpha_d}  - \sum_{i=0}^{x_{nD}-1} (\alpha_D + i)^{-2} }
 \end{bmatrix}
\end{equation}
 
 \begin{equation}\label{eq11}
  \bm{H}(\ell(\alpha, \beta)) = 
  {\displaystyle \sum_{n=1}^N}
 \begin{bmatrix}
  {\displaystyle c_{\alpha\beta} - \sum_{i=0}^{(\sum_D x_{nd})-1} (\alpha + i)^{-2}} & {\displaystyle c_{\alpha\beta}} \\
  {\displaystyle c_{\alpha\beta}} & {\displaystyle c_{\alpha\beta} - \sum_{i=0}^{x_{n(D+1)}-1} (\beta + i)^{-2}}
 \end{bmatrix}
\end{equation}

where the constant terms $ {\displaystyle c_{\alpha_d} = \sum_{i=0}^{(\sum_D x_{nd})-1} \left( \sum_{d=1}^D \alpha_d + i \right)^{-2} }$ and ${\displaystyle c_{\alpha\beta} = \sum_{i=0}^{(\sum_{D+1} x_{nd})-1} (\alpha + \beta + i)^{-2} }$.

The following formulae were used to invert the Hessian efficiently \citealt{minka2000}.

\begin{align*}
 \bm{H} = \text{diag}(\bm{h}) + \bm{11}^\top c
\end{align*}

\begin{equation}\label{eq12}
 \bm{H}^{-1} = \text{diag}(\bm{h}^{-1}) - \frac{\bm{h}^{-1}(\bm{h}^{-1})^\top}{c^{-1}+ \sum_{d=1}^{D} h_d^{-1}}
\end{equation}

where the diag($\cdot$) function places a given vector on the diagonal of a matrix of appropriate dimension, $\bm{h}$ is a column vector containing the non-constant terms from the diagonal of the Hessian, $\bm{1}$ is a column vector of ones of the appropriate dimension, and $c$ is the constant term.  The conditions for which this optimization process maintains convexity required for convergence can be found in Appendix 1. Using \ref{eq7} to perform Newton-Raphson steps during parameter estimation, we considered several computational methods for calculating the gradient and Hessian to compare their accuracy and runtime performance.

\subsubsection{Vectorized MLE}

We first explored the standard approach of exactly calculating the above gradient and Hessian equations using Python Numpy's vectorized operations to improve computational efficiency.  This includes Single-Instruction Multiple Data (SIMD) operations using Streaming SIMD Extensions 2 (SSE2) and intrinsic optimizations of data structures for vector and matrix operations.

\subsubsection{Approximate MLE}

Next, we examined finite limit approximations of the repetitive sums in \ref{eq6} to avoid the computational cost associated with an increasing number of parameters, observations, and data draws.  For this approach, we attempted to approximate the gradient and Hessian finite sums by fitting functions to generalized formulae of this structure:

\begin{equation}\label{eq13}
	f(\theta, N) = \sum_{n=0}^N \frac{1}{\theta + n} = \frac{1}{\theta} + \sum_{n=1}^N \frac{1}{\theta + n}
\end{equation}

\begin{equation}\label{eq14}
	f(\theta, N) = \sum_{n=0}^N \frac{1}{(\theta + n)^2} = \frac{1}{\theta^2} + \sum_{n=1}^N \frac{1}{(\theta + n)^2}
\end{equation}

for positive real $\theta$ and non-negative integer $N$. These formulae, ignoring the term where $n=0$, are extensions of the finite limits for the Harmonic and Geometric series, respectively.  Finite approximations of the divergent Harmonic series and convergent Geometric series have been extensively studied (\citealt{sofo2018,merlini2006}).  However, approximations of these series that include a second parameter in the denominator have, to our knowledge, not been published.  \par

Starting with the known approximation of the trivial case where $\theta \to 0$, manual function fitting was performed by trial and error using the R function "nls" for assistance.  The resulting approximations were then used to compute the finite sums in \ref{eq6} for the Newton-Raphson procedure.  Detailed explanations of the limit approximations are available in Supplementary File 1.

\subsubsection{Sklar's MLE}
Sklar (2014) outlines an approach to reduce the computational cost of the repetitive sums in \ref{eq6} by introducing new data structures.  Here, we extend his approach to the BLM distribution and compare it to the vectorized and approximate MLE approaches.

As noted above, to compute \ref{eq6} naively would involve:

\begin{itemize}
 \item For terms involving $\alpha_d$ and $\beta$, which depend on the count at position $n, d$, repeating sums for each category $d$ less than that of the max value in row $n$ of the data matrix.
 \item For terms involving $\sum_D \alpha_d$, $\alpha$, and $\alpha + \beta$, which depend on the partial or complete row sum of row $n$, repeating sums for each row less than that of the max row sum.
\end{itemize}

To circumvent this, Sklar creates two data structures to hold the multiplicative constants to replace these repetitive sums. Here, due to the increased complexity of the BLM distribution, we need three such data structures. First, we define $Z_D$, $Z_{D+1}$, and $Z_{\text{max}}$ as follows,

\begin{align*}
 Z_D = \underset{n}{\text{max }} \sum_{d=1}^{D} x_{nd} \\
 Z_{D+1} = \underset{n}{\text{max }} \sum_{d=1}^{D+1} x_{nd} \\
 Z_{\text{max}} = \text{max }(Z_D, Z_{D+1})
\end{align*}

The first data structure is a matrix $\underset{(D+1)\times Z_{\text{max}}}{\bm{U}}$, which, for each category $d \in 1 \dots D+1$, counts the number of rows where the count in column $d$ of $\bm{X}$ exceeds the integer $z$, where $z$ is the zero-based column index of $\bm{U}$, $z = 0 \dots Z_{\text{max}}-1$. \par

The second data structure is a vector $\bm{v}^D$ of length $Z_D$, which, for each category $d \in 1 \dots D$, counts the number of rows where the row sum exceeds $z$, where $z$ is the zero-based index of $\bm{v}$, $z = 0 \dots Z_D-1$. \par

The third data structure is another vector $\bm{v}^{D+1}$ of length $Z_{D+1}$, which, for each category $d \in 1 \dots D+1$, counts the number of rows where the row sum exceeds $z$, where $z$ is the zero-based index of $\bm{v}^{D+1}$, $z = 0 \dots Z_{D+1} - 1$.  \par

Note that the dimensionality of the matrix $\bm{U}$ is such that many of its values on the right hand side will be zero using this definition; this is for notational convenience in the derivations.  In practice, the data structure for $\bm{U}$ can be stored efficiently as a ragged array containing only non-zero elements.

More formally:

\begin{align*}
 u_{dz} = \sum_{n=1}^{N} \mathbb{I}\left(x_{nd}>z\right) \\
 v^D_{z} = \sum_{n=1}^N \mathbb{I}\left(\sum_{d=1}^{D} x_{nd}>z\right) \\
 v^{D+1}_{z} = \sum_{n=1}^N \mathbb{I}\left(\sum_{d=1}^{D+1} x_{nd}>z\right)
\end{align*}

Where $\mathbb{I}$ is the indicator function:

\[
 \mathbb{I}(\cdot) \coloneqq 
 \begin{cases}
    1 & \text{if true} \\
    0 & \text{otherwise}
 \end{cases}
\]

This simplifies \ref{eq6}:

\begin{equation}\label{eq15}
 \begin{split}
 \ell(\bm{\theta}) \sim \sum_{z=0}^{Z_D-1} \left[ \sum_{d=1}^D u_{dz}\ln(\alpha_d + z) + v_z^D \ln(\alpha + z) - v^D_z \ln(\sum_{d=1}^{D} \alpha_d + z) \right] \\
 + \sum_{z=0}^{Z_{D+1}-1} \left[ u_{(D+1)z} \ln(\beta + z) -  v^{D+1}_z\ln(\alpha+\beta+z) \right]
 \end{split}
\end{equation}

The gradient and Hessian of (16) are then,

\begin{align*}
 g_{\alpha_d} = {\displaystyle \sum_{z=0}^{Z_D-1} \left[ u_{dz} (\alpha_d+z)^{-1} - v_z^D (\sum_{d=1}^D \alpha_d +z)^{-1} \right] }
\end{align*}

\begin{equation}\label{eq16}
 g_{\alpha} = {\displaystyle \sum_{z=0}^{Z_D-1} v_z^D (\alpha + z)^{-1} - \sum_{z=0}^{Z_{D+1}-1} v_z^{D+1} (\alpha+\beta+z)^{-1}}
\end{equation}

\begin{align*}
 g_\beta = {\displaystyle \sum_{z=0}^{Z_{D+1}-1} \left[ u_{(D+1)z} (\beta+z)^{-1} - v_z^{D+1} (\alpha+\beta+z)^{-1} \right] }
\end{align*}

 \begin{equation}\label{eq17}
 \bm{H}(\ell(\alpha_1, \dots, \alpha_D)) =
 {\displaystyle \sum_{z=0}^{Z_{D}-1} }
 \begin{bmatrix}
  {\displaystyle  c_{\alpha_d} - u_{dz} (\alpha_d + z)^{-2} } & \dots & {\displaystyle c_{\alpha_d} } \\
  \vdots & \ddots & \vdots \\
  {\displaystyle c_{\alpha_d}} &  \dots & {\displaystyle c_{\alpha_d} - u_{dz} (\alpha_d + z)^{-2}}
 \end{bmatrix}
\end{equation}

 \begin{equation}\label{eq18}
 \bm{H}(\ell(\alpha, \beta)) =
 \sum_{z=0}^{Z_{D+1}-1}
 \begin{bmatrix}
  {\displaystyle  c_{\alpha\beta} - \sum_{z=0}^{Z_D-1} v_z^D (\alpha + z)^{-2}} & {\displaystyle c_{\alpha\beta}} \\
  {\displaystyle c_{\alpha\beta}} & {\displaystyle c_{\alpha\beta} - u_{(D+1)z}(\beta + z)^{-2} }
 \end{bmatrix}
\end{equation}

where the constants $c_{\alpha_d} = v_{z}^D(\sum_{d=1}^{D} \alpha_d + z)^{-2}$ and $c_{\alpha\beta} = v_z^{D+1}(\alpha + \beta + z)^{-2}$.

\subsection{MLE Runtime Analysis}
Theoretical and empirical runtimes for the MLE Newton-Raphson procedure were assessed for the DM and BLM distributions using the three computational methods listed above: vectorized, approximate, and Sklar. For empirical runtime assessment, the number of observations $N$, the number of data draws from each observation vector $M = \sum_{D+1} x_d$, and the number of categories $D+1$, were each varied while holding the others constant.  Wall-clock time for a single Newton-Raphson MLE step was then computed for each of the above algorithms for the DM and BLM distributions.  Five independent bootstraps were performed for each experiment to account for variance due to hardware performance.  All runtime experiments were performed on Ubuntu Linux 18.04 using Python 3.7.3, Numpy 1.18.1, an 18-core Intel i9-9980XE Skylake 3.0 GHz processor, 64 GB DDR4 3600 SDRAM, and an M.2 2280 NVMe 1.3 V-NAND solid state drive.

\subsection{Classification Performance Benchmarking}
We aimed to characterize the effectiveness of the BLM distribution for classifying labelled data, commonly known as supervised classification.  In the following sections, we compare the BLM to the DM and multinomial distributions using both simulated data and real-life, gold standard datasets, all of which have known class labels.  Simulated data allows for fine control over the difficulty of the classification task and demonstrates for which datasets these distributions are effective.  Gold standard datasets offer additional value over simulated data, as they often contain noise unrelated to the classification task and are therefore more representative of how each distribution will perform in general. \par

In the following experiments, we compare standard multinomial naive Bayes and two MLE-based strategies for performing the classification task.  Each of these strategies uses different information from the training and test data, however all methods involve maximizing the likelihood function in some way. For the standard multinomial and the first of the MLE-based classification strategies, we construct a test matrix of counts for each feature $\underset{N \times (D+1)}{\bm{T}}$ and a training matrix with columns representing each class, $c \in C$, on the unit simplex $\underset{(D+1) \times C}{\bm{X}}$.  Classification of multiple test observations is then performed by computing the matrix inner product and taking the argmax for each row in the resulting matrix, which corresponds to the assigned class for each test observation. A formal explanation of this process is available in Supplementary File 1. The second MLE-based approach involves explicit calculation of the marginal likelihood for each class and does not involve calculating the matrix inner product. Below, we provide additional details about these methods, referred to here as multinomial naive Bayes, Dirichlet and Beta-Liouville multinomial naive Bayes, and marginal likelihood classification, respectively. \par

\textbf{Multinomial Naive Bayes} \\
Parameters on the unit simplex are determined by simple division for multinomial naive Bayes. In this approach, observations $n \in N$ within the training data are summed to produce a single count vector, and the counts are divided by the vector sum to produce the training parameters, equivalent to the MLEs of the multinomial, $\bm{\hat{p}}$.  The resulting trained parameters are then used to classify test observations by matrix inner product.

\begin{equation}\label{eq19}
	\hat{p}_d = \frac{\sum_N x_{nd}}{\sum_N \sum_{D+1} x_{nd}}, \text{ for } d=1 \dots D+1
\end{equation}

\textbf{Dirichlet and Beta-Liouville Multinomial Naive Bayes} \\
In this approach, the training data are used to perform Newton-Raphson MLE separately for each class, and the resulting $D+1$ and $D+2$ MLEs are used to calculate the $D+1$ training parameters for the DM and BLM distributions, respectively.  The resulting trained parameters are then used to classify test observations by matrix inner product. \\

DM Distribution: \\
\begin{equation}\label{eq20}
	\hat{p}_d = \frac{\hat{\alpha}_d}{\sum_{D+1} \hat{\alpha}_d}, \text{ for } d=1 \dots D+1 
\end{equation}

BLM Distribution: \\
\begin{equation}\label{eq21}
	\hat{p}_d = \frac{\hat{\alpha}}{\hat{\alpha} + \hat{\beta}} \frac{\hat{\alpha}_d}{\sum_{D+1} \hat{\alpha}_d}, \text{ for } d=1 \dots D
\end{equation}

\begin{align*}
	\hat{p}_{D+1} = \frac{\hat{\beta}}{\hat{\alpha} + \hat{\beta}}
\end{align*}

Where $\hat{\alpha}_d$, $\hat{\alpha}$, and $\hat{\beta}$ are the MLEs.

\textbf{Marginal Likelihood Classification} \\
Like the previous approach, the training data are used to calculate MLE parameter estimates for $D+1$ and $D+2$ parameters for the DM and BLM distributions, separately for each class.  However, for classification, the marginal likelihood function $P(\bm{x} | \bm{\theta})$ is evaluated separately for each test observation vector $\bm{x}$ and class parameter MLE estimates $\bm{\theta}_c, c \in C$.  The test observation is then assigned to the class with the maximum likelihood.

DM Distribution: \\
\begin{equation}\label{eq22}
	\displaystyle{ \ln(P(\bm{x} | \bm{\theta}_c)) = \sum_{i=0}^{(\sum_{D+1} x_d) - 1} \ln(1 + i)  - \sum_{i=0}^{x_d - 1} \ln(1 + i) + \sum_{d=1}^{D+1} \sum_{i=0}^{x_d - 1} \ln(\hat{\alpha}_d + i)  - \sum_{i=0}^{(\sum_{D+1} x_d) - 1} \ln(\sum_{d=1}^{D+1} \hat{\alpha}_d + i) }
\end{equation}

BLM Distribution: \\
\begin{equation}\label{eq23}
    \begin{split}
        \ln(P(\bm{x} | \bm{\theta}_c)) = {\displaystyle \sum_{i=0}^{(\sum_{D+1} x_d)-1} \ln(1 + i) - \sum_{d=1}^{D+1} \sum_{i=0}^{x_d - 1} \ln(1 + i) + \sum_{d=1}^{D} \sum_{i=0}^{x_{d}-1} \ln(\hat{\alpha}_d + i) - \sum_{i=0}^{\left(\sum_D x_{d}\right) -1} \ln\left(\sum_{d=1}^{D} \hat{\alpha}_d + i\right) } \\
        {\displaystyle + \sum_{i=0}^{(\sum_D x_{d})-1} \ln( \hat{\alpha} + i) + \sum_{i=0}^{\left(x_{D+1}\right)-1} \ln(\hat{\beta} + i) - \sum_{i=0}^{\left(\sum_{D+1} x_{d}\right)-1} \ln(\hat{\alpha} + \hat{\beta} + i) }
    \end{split}
\end{equation}

Finally, Laplace additive smoothing was applied to all of the above methods such that all categories had non-zero training parameter probability even if they were not observed in the training data (\citealt{he2007}).

\subsubsection{Classification Power Analysis on Simulated Data}
A power analysis was performed for the DM and BLM distributions to assess and compare classification accuracy on simulated data.  Data were simulated to achieve different levels of overlap and sampling for 4 distinct classes. Benchmarking on these data involved assessing the accuracy of classification back to the known class labels using standard performance measures, including precision, recall, specificity, and F1 score. Data were generated as follows (Figure 1).

\begin{itemize}
	\item We set the number of categories, $D+1$, within each class to be 100. These are categories of a generating multinomial with parameters $\bm{p}=\{p_1, \dots, p_{100}\}$.  
	\item We chose fixed, equidistant categories to center each class: category 1 for class 1, category 33 for class 2, category 66 for class 3, and category 100 for class 4. These centers represent the probability density mode of the associated multinomial for each class (Figure 1A).
	\item The vector $\bm{p}$ was generated as follows for each class: 1) We used the central category of the class as a mean of a normal distribution and varied the standard deviation to create more or less overlap between class probability densities (Figure 1B).  Five standard deviation values were explored: 10, 22, 35, 48, and 60. 2) We randomly sampled 2000 draws from these normal distributions associated with each class. For example, sampling from the normal distribution with mean 33 and standard deviation 10 associated with class 2 results in data points spread, for the most part, between category 16 and category 49 with a mode around the 33 mean. We created an empirical distribution by rounding the sampled numbers to integers and counting the integers with the same value. For example, we could count the number of integers that were equal to 30 (category 30). 3) For each class we divided the totals per category by the total number of draws to construct the probability vector $\bm{p}$ of the generating multinomials. Large standard deviation values result in more overlap between the generating multinomials per class, while small standard deviations result in more class separation with concentrated probability density around the mean.
	\item The resulting multinomials with different degree of overlap were used to generate count data for subsequent classification. To assess the impact of sampling on these results we varied the total number, $M$, of samples (data draws) obtained from the generating multinomials.  Values of 15, 136, 258, 379, and 500 were explored for $M$. Small values of $M$ result in sparse simulated count vectors, while large values of $M$ result in dense count vectors that better inform the maximum likelihood estimation process. 
	\item Utilizing the same generating multinomial distributions, we also varied the number of count vectors $N$ (data observations) generated for each class.  Values of 2, 26, 51, 76, and 100 were explored for $N$. Larger values of $N$ should result in better parameter estimates and therefore more accurate classification.
\end{itemize}

These datasets were preprocessed as outlined in Supplementary File 1 and classified using the methods introduced in previous sections.

\begin{figure}[h!]
  \centering
  \includegraphics[width=\textwidth]{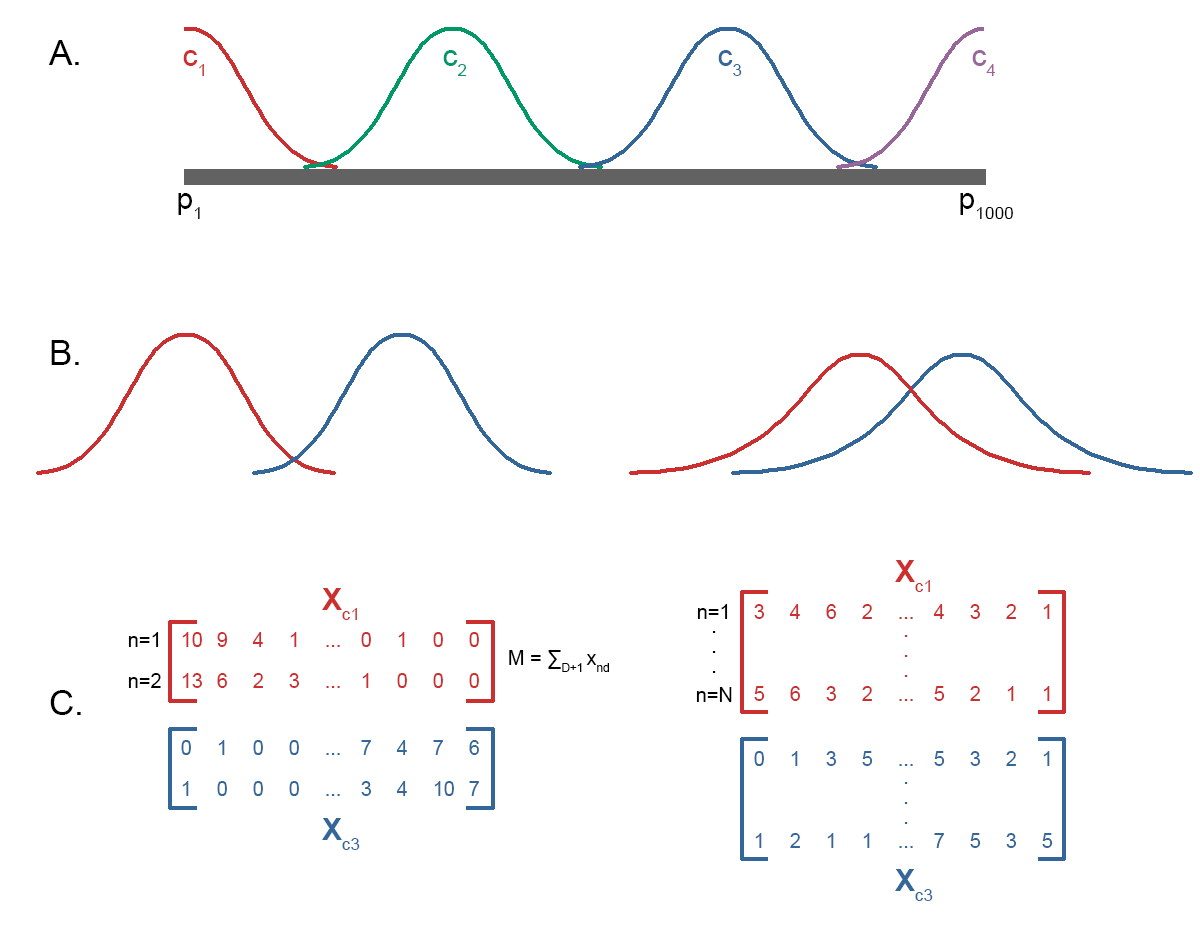}
  \caption{\textbf{Data were simulated for 4 discrete classes according to normal distributions evenly spaced along a vector of 100 categories.}  A) Each class had normal probability density with mean equal to a category varying equidistantly from $p_1$ to $p_{100}$. B) For each class, the standard deviation of the normal distribution was varied, $\sigma=\{10, 22, 35, 48, 60\}$, with higher values creating more overlap between classes. C) Using the resulting multinomial parameters, $M=\{15, 136, 258, 379, 500\}$ counts were randomly sampled for $N=\{2, 26, 51, 76, 100\}$ independent data vectors for each class.  These counts were then classified according to previously stated methods and used to calculate accuracy measures.}
  \label{fig:fig1}
\end{figure}

Finally, we aimed to determine if the BLM distribution performed better on count data derived from a BLM distribution as opposed to multinomial data.  To achieve this, multinomial generating parameters were simulated as above and were used in a Markov-chain Monte Carlo (MCMC) process to produce BLM data with target distribution, \par

\begin{align}\label{eq24}
	P(\bm{x}_c|\bm{\theta}_c) = \frac{P(\bm{\theta}_c|\bm{x}_c)P(\bm{x}_c)}{P(\bm{\theta}_c)}
\end{align}

where $\bm{x_c}$ is a count data vector, and $\bm{\theta_c}$ are the fixed, generating parameters, for class $c$.  A formal description of the MCMC data simulation process can be found in Supplementary File 1. MCMC simulation was performed separately for each class using a Metropolis-Hastings sampling strategy with a multinomial proposal, $P(\bm{x})$, and acceptance-rejection probability,

\begin{equation}\label{eq25}
 \alpha = \text{min} \left\{1,\frac{P(\bm{x}_{i+1}|\bm{\theta}_c)/P(\bm{x}_{i+1})}{P(\bm{x}_i|\bm{\theta}_c)/P(\bm{x}_i)}\right\}
\end{equation}

where $P(\bm{x}_{i+1}|\bm{\theta}_c)$ is the likelihood of $\bm{x}_{i+1}$ under the BLM and $P(\bm{x}_{i+1})$ is the likelihood of the data under the proposal. The data generation process for both multinomial and BLM data was independently replicated 5 times for each set of variables, and the resulting datasets were classified using the same procedures introduced in the previous sections.

\subsubsection{Classification on Gold Standard Datasets}
Performance on real datasets was determined using four gold standard datasets from the text-based natural language processing domain (\citealt{cardoso2007-1}) described below:

\begin{itemize}
	\item \textbf{20 Newsgroups}: an English text collection of 18,821 newsgroup documents categorized into 20 classes with a relatively even class distribution. The dataset used in this experiment was split within each class into 11,293 training and 7,528 testing documents.
	\item \textbf{Reuters-21578 (R8)}: an English text collection of 7,674 newswire documents from the Reuters Ltd. group in 1987.  Only the 10 most frequent classes were used for this dataset, and only observations belonging to a single, unique class label were included.  This resulted in 8 classes with more than one document, hence the name R8.  This dataset has a heavily skewed class distribution and was split within each class into 5,485 training and 2,189 test documents.
	\item \textbf{Cade12}: a Brazilian Portuguese text collection of 40,983 webpages from the CADE web directory.  Documents are categorized into 12 classes with a skewed class distribution.  This dataset was split within each class into 27,322 training and 13,661 test documents.
	\item \textbf{WebKB}: an English text collection of 4,199 university webpages from computer science departments participating in the World Wide Knowledge Base project in 1997.  Documents are categorized into 4 classes with a moderately skewed class distribution.  This dataset was split within each class into 2,803 training and 1,396 test documents.
\end{itemize}

\section{Results}
MLE runtimes for the BLM distribution were comparable to the DM distribution but with a small increase in runtime due to computation of additional terms in the likelihood function and its derivatives (Figure 2).  Performance of the BLM distribution on simulated data was nearly identical to the DM distribution's performance. However, the BLM distribution was not as accurate on very sparse data with few observations, $N$, and few data draws, $M$, suggesting that the BLM distribution requires more data density than the DM to produce accurate MLEs. Finally, the multinomial distribution performed most consistently on gold standard data and performed best on two of the four datasets examined.  However, the DM and BLM distributions were able to outperform the multinomial in accuracy by a small margin on the other datasets.  Additional details about the results of each experiment are provided below.

\subsection{MLE Runtime Results}
Overall, the vectorized method was the most computationally efficient, particularly on data with a large number of parameters and dense vectors (a large number of data draws).  The approximate method performed optimally for high values of data draws and low values of data observations.  Sklar's method performed optimally with a large number of data observations and a small number of parameters and data draws; however, the runtime of Sklar's approach scaled poorly relative to the vectorized and approximate methods as the number of parameters increased.  Of particular note are the middle and right panels in Figure 2, highlighting impact of an increased number of parameters and number of draws, respectively, on runtime. The figure shows that Sklar's method scaled exponentially as parameter count increased, reaching almost 60 seconds per Newton-Raphson step at a moderate parameter count of 5,000. In contrast, the vectorized method scaled linearly, requiring only seconds per step, and the approximate method scaled nearly log-linearly with the number of data draws.  This suggests that the vectorized and approximate approaches will perform most efficiently for a majority of real-world problems, for which a large number of parameters must be estimated.

\begin{figure}[h!]
  \centering
  \includegraphics[width=\textwidth]{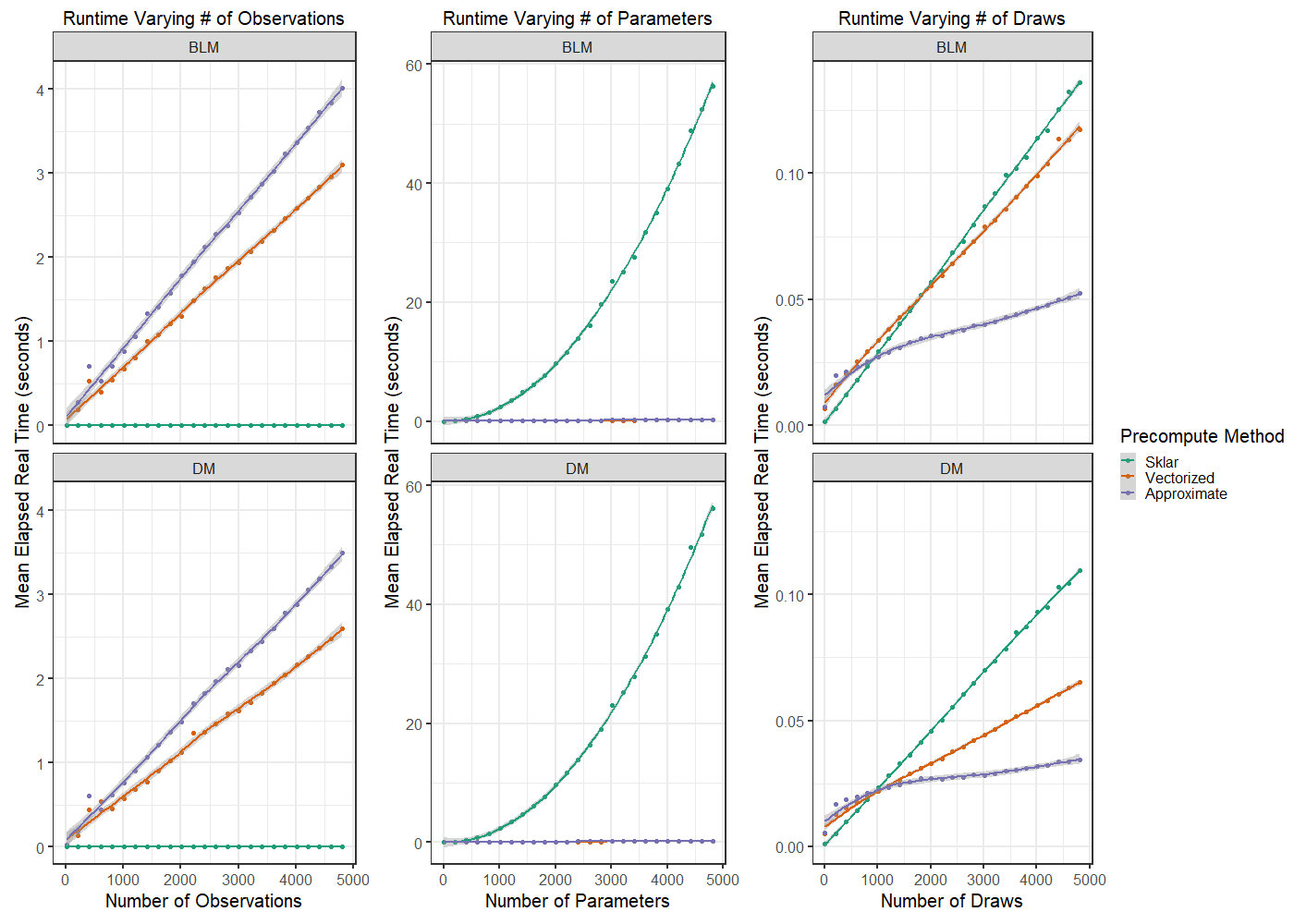}
  \caption{\textbf{Vectorized and approximate methods scale more efficiently than Sklar's method with a large number of parameters and data draws.} Sklar's approach scales more efficiently than the vectorized and approximate methods when the number of observations is large and the number of parameters and data draws are small. The approximate method is the most computationally efficient when the number of data draws is large, as it is capable of scaling log-linearly.  In the above figures, each data point is the mean of 5 independent bootstraps, and a locally estimated scatterplot smoothing (LOESS) line, as implemented in the R ggplot2 package, is shown for each method (\citealt{cleveland1992}). Variance around the LOESS line is shown in gray.}
  \label{fig:fig2}
\end{figure}

\subsection{Classification Results}
Overall, the standard multinomial naive Bayes classifier performed consistently well on both simulated and gold standard data classification tasks.  The DM and BLM MLEs resulted in marginally better classification performance for the 20 Newsgroups and WebKB gold standard datasets but were not as consistent as the multinomial classifier, likely due to the relative difficulty of parameter optimization compared to the simple division used to produce the multinomial parameter estimates.  As expected, all distributions and methods improved in classification performance when more data were available, including both the number of observations $N$ and the number of data draws $M$ within each observation. The following sections explore more precisely how classification performance changed with varying amounts of data and how well these power estimates generalized to real-world data.

\subsubsection{Simulated Data Power Analysis Results}
The following variables affected classification power, ordered from most to least important: 1) difficulty of the classification task (i.e. class distribution overlap), 2) the density of the data vector (data draws), and 3) the number of data observations (Figure 3).  Even for heavily overlapping distributions, median F1 score exceeded 90\% for the matrix multiplication methods using the multinomial, DM, and BLM distributions at 26 observations and 136 data draws (Supplementary Figures 1-2).  The classification scores for classes 1 and 4 were typically higher than those for classes 2 and 3; this was due to classes 1 and 4 having less overlap than classes 2 and 3, resulting in an easier classification task (Figure 1). However, these class differences were resolved with additional training data, particularly for the datasets with less class overlap. \par

The marginal likelihood classification method required more data observations and data draws to achieve a similar F1 score on the same data, with the BLM distribution requiring more data than the DM distribution (Supplementary Figures 3-4).  Finally, data generated from the multinomial distribution did not appear to affect the F1 score when compared to data generated from the BLM distribution for all classification methods and distributions examined (Supplementary Figures 5-6). This suggests that the major differences in classification power were due to differences in the distributions and classification methods but not the statistical origin of the data. Overall, these results demonstrate that the multinomial, DM, and BLM distributions are capable of achieving nearly perfect classification results under ideal conditions, provided enough training data are available.

\begin{figure}[h!]
  \centering
  \includegraphics[width=\textwidth]{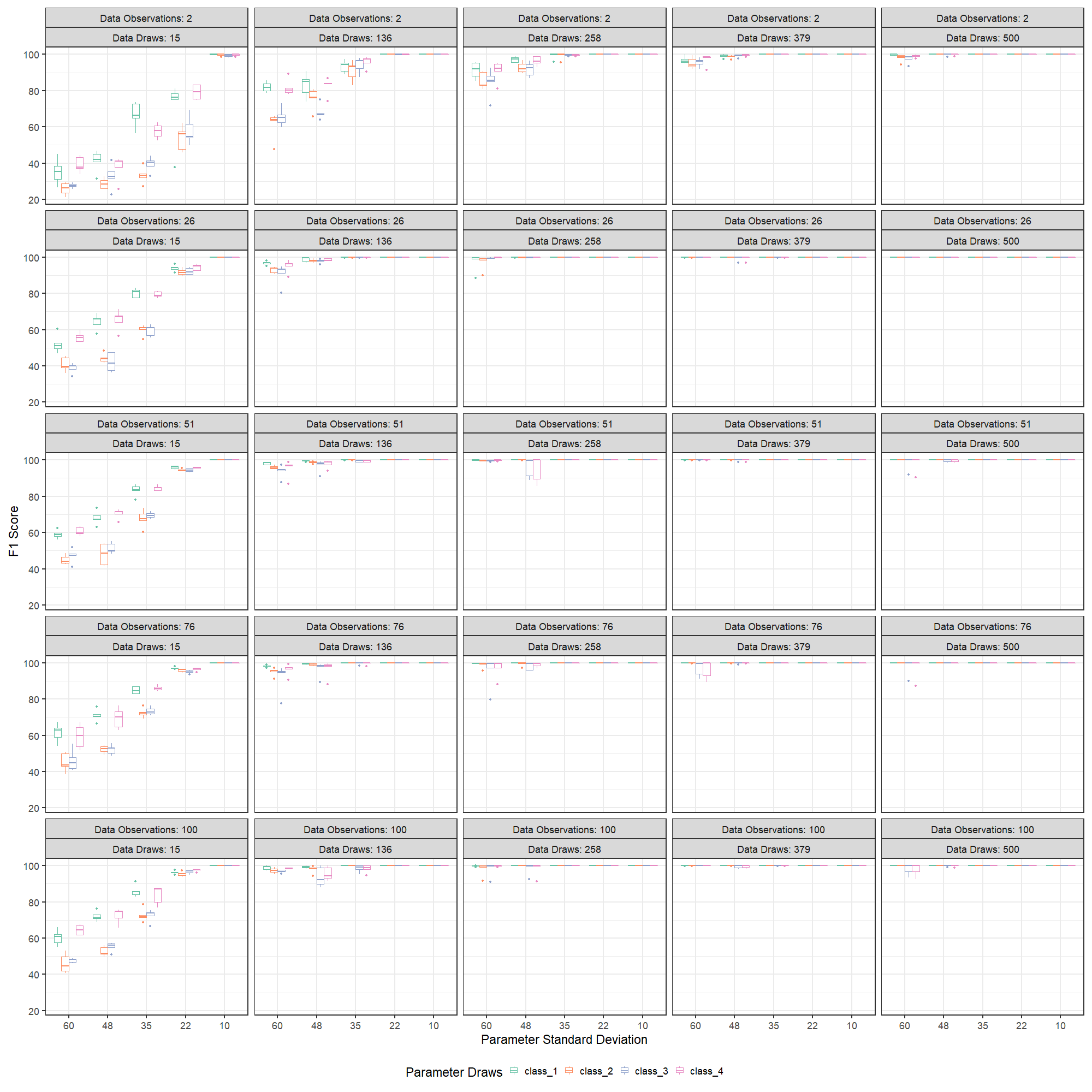}
  \caption{\textbf{Use of the BLM distribution results in high F1 classification scores as data draws and observations increase, even for difficult classification tasks.} Parameter standard deviation is a measure of classification difficulty: higher values result in more overlapping class distributions. Data draws (the density of the count vectors) has a greater affect on F1 score than data observations, however higher values of both result in higher F1 scores.  A separate boxplot is shown for each simulated class distribution (classes 1-4).  The boxplots summarize the F1 score quartiles for 5 independent datasets for each combination of variables.}
  \label{fig:fig2}
\end{figure}

\subsubsection{Gold Standard Data Results}
Each of the DM, BLM, and multinomial distributions had the highest accuracy and F1 scores on two of the four gold standard datasets (Table \ref{tab2}).  For all datasets and distributions, the naive Bayes classification methods using the matrix inner product produced the highest performance metrics.  For the DM and BLM MLEs, the approximate compute method resulted in marginally better performance metrics than the exact methods (vectorized and Sklar) for the 20 Newsgroups and R8 datasets.  However, the approximate method was not consistent and resulted in markedly lower accuracy and F1 scores for the Cade12 dataset.  The lower accuracy of the approximate method on the Cade12 dataset could be due to the MLEs converging on large parameter values, for which the finite sum approximations diverge from exact values.  This limitation of the approximate method is discussed further in the following section. The multinomial distribution with a standard naive Bayes classifier performed consistently well across all datasets, while the DM and BLM MLEs were less consistent overall.  \par

Of particular interest, the Cade12 dataset had markedly lower F1 scores (around 7\% lower) for the MLEs than the multinomial distribution across all classification and compute methods.  In contrast, the MLEs performed almost as well or better than the multinomial distribution for the remaining datasets to within nearly 1\% F1 score. This suggests that the Cade12 dataset poses challenges for the accurate calculation of the MLEs, and it is well-known that the Cade12 data are more complex and difficult to classify than the other datasets evaluated here (\citealt{cardoso2007-2}).  Additionally, the Cade12 dataset was the only dataset for which the approximate compute method and marginal likelihood classifiers produced extremely low F1 scores relative to the exact compute methods (vectorized/Sklar) using the naive Bayes classifier.  These results are suggestive of numerical overflow during computation or that the Newton-Raphson method is traversing a difficult optimization landscape.  In the discussion of these results, we outline additional criteria for which the approximate method breaks down that could have contributed to the low F1 scores.  \par

Finally, we note that because classification results were calculated across all classes within each dataset, the F1 score was identical to both precision and recall; therefore precision and recall are not shown. 

\begin{table}[h!]
\centering
\caption{\textbf{The DM, BLM, and multinomial distributions each performed best on two of the four gold standard datasets.} The best-performing combination of distribution and method are shown in bold for each dataset.  In all datasets, the standard multinomial naive Bayes classifier performed consistently well, while the MLE-based methods were less consistent across datasets.  The approximate compute method appeared to result in better MLEs for the 20 Newsgroups and R8 datasets, perhaps by preventing computational overflow.  Percent accuracy and F1 score are presented for each combination of methods and datasets examined.}
\begin{adjustbox}{center}
\begin{tabular}{|*{11}{c|}}
\hline
\multirow{2}{*}{\textbf{Distribution}} & \multirow{2}{*}{\textbf{Classifier}} & \multirow{2}{*}{\textbf{Compute Method}} & \multicolumn{2}{|c}{\textbf{20 Newsgroups}} & \multicolumn{2}{|c}{\textbf{Reuters-21578 (R8)}} & \multicolumn{2}{|c}{\textbf{Cade12}} & \multicolumn{2}{|c|}{\textbf{WebKB}} \\ \cline{4-11}
 & & & Accuracy & F1 & Accuracy & F1 & Accuracy & F1 & Accuracy & F1 \\ \hline
\multirow{4}{*}{DM} & \multirow{2}{*}{Naive Bayes} & Vectorized/Sklar & 98.2 & 82.2 & 98.7 & 94.8 & 91.7 & 50.3 & 93.0 & 86.0 \\ \cline{3-11}
 & & Approximate & \textbf{98.3} & \textbf{82.9} & \textbf{99.2} & \textbf{96.7} & 86.6 & 19.6 & 93.1 & 86.1 \\ \cline{2-11}
 & \multirow{2}{*}{Marginal} & Vectorized/Sklar & 97.8 & 78.5 & 97.9 & 91.4 & 88.8 & 32.9 & 92.9 & 85.8 \\ \cline{3-11}
 & & Approximate & 98.2 & 81.6 & 98.8 & 95.2 & 86.1 & 16.8 & 92.9 & 85.8 \\ \hline
 \multirow{4}{*}{BLM} & \multirow{2}{*}{Naive Bayes} & Vectorized/Sklar & 98.2 & 82.2 & 98.7 & 94.8 & 91.7 & 50.3 & 93.0 & 86.0 \\ \cline{3-11}
 & & Approximate & 97.8 & 77.7 & \textbf{99.2} & \textbf{96.7} & 86.6 & 19.7 & 92.2 & 84.4 \\ \cline{2-11}
 & \multirow{2}{*}{Marginal} & Vectorized/Sklar & 97.8 & 78.1 & 97.7 & 90.6 & 86.4 & 18.6 & \textbf{93.2} & \textbf{86.3} \\ \cline{3-11}
 & & Approximate & 97.7 & 76.6 & 98.8 & 95.2 & 85.8 & 15.0 & 90.4 & 80.7 \\ \hline
 Multinomial & Naive Bayes & Multinomial & 98.2 & 82.5 & \textbf{99.2} & \textbf{96.7} & \textbf{93.0} & \textbf{57.9} & 92.0 & 84.1 \\ \hline
\end{tabular}
\end{adjustbox}
\label{tab2}
\end{table}

\section{Discussion}
Our results show that efficient methods exist for computation of MLEs for the Dirichlet Multinomial (DM) and Beta-Liouville Multinomial (BLM) distributions.  For datasets where the number of categories (features $D+1$) to be estimated is small but the data matrix (observations $N$ and count density $M$) is large, Sklar's computational approach would be the most efficient.  However, it is common in natural language processing classification tasks to have many features (often more than 1,000) and sparse data available for MLE computation.  In this more common case, the vectorized computation and the limit approximation approaches explored here would be far superior for computational efficiency.  \par 

Our results also indicate that the BLM distribution seems to perform equally as well as the DM distribution on both simulated and gold standard data, outperforming the DM distribution for certain datasets like WebKB (Figure 3, Table 2).  Particularly, the BLM distribution performs well when sufficient data are present for optimization of the BLM parameters and when the class densities are mildly to moderately non-overlapping.  For datasets where the class distributions are heavily overlapping and data are sparse, such as in the Cade12 dataset, the DM and BLM MLEs performed worse than the naive estimates produced by the standard multinomial naive Bayes procedure.  This suggests that the optimization landscape for such datasets is not sufficiently easy to navigate for the Newton-Raphson procedure, resulting in poor MLEs.  However, we note that the BLM and DM distributions may perform better under different circumstances like unsupervised classification, where class labels are not fixed, as has been shown by Bouguila (\citeyear{bouguila2008}).  The results observed here for the Cade12 dataset may be a limitation of the DM and BLM distributions in strictly supervised classification tasks. \par 

An interesting result was the marginal gain in performance using the finite limit approximations, leading to a 0.7\% F1 gain for the DM naive Bayes on the 20 Newsgroups dataset compared to all other methods and a 1.9\% F1 gain for the BLM and DM naive Bayes on the R8 dataset compared to the MLE naive Bayes methods.  This suggests that the approximated values produced a more rapidly-converging Newton-Raphson optimization landscape than the exact methods (vectorized/Sklar). It is also possible that the approximations helped to avoid numerical overflow in cases where $\theta \to 0$, which results in large values for the first iteration of the finite sums. Additionally, our approximations seemed to be accurate for small parameter values ($\theta$); the parameter MLEs tended toward smaller values for the BLM and DM distributions on the 20 Newsgroups, R8, and WebKB datasets (data not shown).  However, our approximations appeared to be inaccurate as parameter MLEs reached large values ($\theta > 1000$), as was reflected in the poor performance metrics for the DM and BLM distributions on the Cade12 dataset.  If more exact approximations were found for these finite sums, they could result in increased computational efficiency and more rapid convergence for Newton-Raphson MLEs. Therefore, while our approximations produced an interesting result, further work must be done to derive more exact equations before they are employed for general use. \par

Though the BLM MLEs were successful on two of the four gold standard datasets, it became clear during our research that the BLM distribution has several limitations during Newton-Raphson optimization that must be avoided for its successful use in this context.  We observed two types of pathological behavior during parameter optimization that hadn't been previously described, to the best of our knowledge.  First, the Hessian matrix of the BLM requires that values for $\alpha_1, \alpha, \beta$ meet certain conditions for the matrix to be negative definite (Appendix I).  If the Hessian is not negative definite, then convergence of the Newton-Raphson procedure is not gauranteed.  We solved this problem by checking these conditions before each Newton-Raphson step and adding small values to pathological parameters, such that the Hessian matrix was gauranteed to be negative definite.  Second, even when convergence was gauranteed, we encountered certain datasets for which $\alpha, \beta$ had rapidly increasing parameter values, resulting in eventual numerical overflow.  In these cases, the ratio of $\frac{\beta}{\alpha}$ remained constant while the parameter estimates increased, suggesting the existence of local ridge optima in the landscape. This could be due to overparameterization of these particular datasets, suggesting that only $D+1$ parameters are needed to model the data, while the BLM has $D+2$ parameters.  In these cases, it is recommended that the DM distribution be used instead of the BLM.  The code accompanying this work contains examples of checking for these pathological behaviors and potential solutions to ensure convergent and accurate MLEs for the BLM distribution. 

We acknowledge that the scope of this work is limited to naive Bayes classifiers, and that the power and gold standard data analyses presented here are not comprehensive.  There may be other classifiers and datasets that demonstrate improved results for the DM and BLM distributions.  Likewise, there may be datasets for which the DM and BLM distributions produce poor results regardless of the classification method used.  Finally, while the DM and BLM distributions may produce marginal increases in accuracy for certain datasets, the time cost for parameter optimization may outweigh any increase in accuracy provided by their use.  Given that the multinomial naive Bayes classifier performed consistently well and required a fraction of the computational cost, it may be the superior choice for supervised classification tasks. \par

Future work might explore additional optimization methods other than Newton-Raphson, additional classifiers other than naive Bayes, and broader classification tasks like unsupervised or streaming classification.  Likewise, further exploration of finite sum approximations for the modified geometric and harmonic series presented here could be exciting, since the general structure of these equations appear often in gradient and Hessian calculations for a variety of distributions. Finally, application of the DM and BLM distributions could be explored in areas other than the bag-of-words natural language processing classification task, such as problems in bioinformatics and ecology.

\section{Acknowledgements and Additional Information}

\subsection{Funding Details}
This work was supported by the United States Department of Agriculture Animal and Plant Health Inspection Service under the National Bio- and Agrodefense Facility Scientist Training Program; and the National Science Foundation under grant DGE-1450032.

\subsection{Disclosure Statement}
The authors have no conflicts of interest to disclose.

\subsection{Supplementary Files}
Supplementary File 1 includes all supplementary figures and additional information regarding limit approximations, classification methods, and gold standard data preprocessing.

\subsection{Data Availability}
All code used to generate the data and figures presented here is available online at \url{https://github.com/lakinsm/fast-mle-multinomials} and is preserved in its current state under the Publication release.

\clearpage

\bibliographystyle{apalike}
\bibliography{LakinAbdo_FastMLEBetaLiouville_Manuscript} 

\clearpage

\section{Appendix I}

It is sufficient to show that the Hessian matrix, associated with the likelihood function, is negative definite to guarantee a unique global optimum and attain the BLM MLEs. Here we show the form of the Hessian matrix and provide conditions required for it to be negative definite. Continuing from the BLM Hessian described in equations \ref{eq10} and \ref{eq11}, the components of the Hessian are,

\begin{align}
  \frac{\delta^2F(\bm{\theta})}{\delta^2\alpha} & = -\sum_{i=0}^{(\sum_{d=1}^Dx_d)-1} \frac{1}{(\alpha + i)^2} + \sum_{i=0}^{(\sum_{d=1}^{D+1}x_d)-1} \frac{1}{(\alpha + \beta + i)^2}\nonumber \\
  \frac{\delta^2F(\bm{\theta})}{\delta^2\beta} & = - \sum_{i=0}^{x_{D+1}-1} \frac{1}{(\beta + i)^2} + \sum_{i=0}^{(\sum_{d=1}^{D+1}x_d)-1} \frac{1}{(\alpha + \beta + i)^2}\nonumber \\
  \frac{\delta^2F(\bm{\theta})}{\delta \alpha \delta \beta} & = \sum_{i=0}^{(\sum_{d=1}^{D+1}x_d)-1} \frac{1}{(\alpha + \beta + i)^2}\nonumber \\
  \frac{\delta^2F(\bm{\theta})}{\delta^2\alpha_d} & = - \sum_{i=0}^{x_d - 1} \frac{1}{(\alpha_d + i)^2} + \sum_{i=0}^{(\sum_{d=1}^Dx_d)} \frac{1}{(\sum_{d=1}^D\alpha_d + i)^2} \nonumber \\
  \frac{\delta^2F(\bm{\theta})}{\delta \alpha_d \delta \alpha_{g}} & =\sum_{i=0}^{(\sum_{d=1}^Dx_d)} \frac{1}{(\sum_{d=1}^D\alpha_d + i)^2} \forall \{d \neq g\} 
\end{align}

Where,

\begin{align}
  0 <  a_{\alpha} & = \sum_{i=0}^{(\sum_{d=1}^Dx_d)-1} \frac{1}{(\alpha + i)^2} \nonumber \\
  0 <  a_{\beta} & = \sum_{i=0}^{x_{D+1}-1} \frac{1}{(\beta + i)^2} \nonumber \\
  0 <  c_1 & = \sum_{i=0}^{(\sum_{d=1}^{D+1}x_d)-1} \frac{1}{(\alpha + \beta + i)^2} \nonumber \\
  0 <  a_{\alpha_d} & = \sum_{i=0}^{x_d - 1} \frac{1}{(\alpha_d + i)^2} \nonumber \\
  0 <  c_2 & = \sum_{i=0}^{(\sum_{d=1}^Dx_d)} \frac{1}{(\sum_{d=1}^D\alpha_d + i)^2}
\end{align}

Then,

\begin{equation}
  \bm{H}(\bm{\theta}) = 
  \begin{pmatrix}
    -a_{\alpha} + c_1 & c_1 & 0 & 0 & 0 & \hdots \\
    c_1 & -a_{\beta} + c_1 & 0 & 0 & 0 & \hdots \\
    0 & 0 & -a_{\alpha_1} + c_2 & c_2 & c_2 & \hdots \\
    0 & 0 & c_2 & -a_{\alpha_2} + c_2 & c_2 & \hdots \\
    \vdots & & \ddots & & \hdots
  \end{pmatrix}
\end{equation}

This is a symmetric matrix and is negative definite if its pivots are negative. Using Gaussian elimination, the above matrix can be reduced to,

\begin{equation}
  \begin{pmatrix}
    - a_{\alpha} + c_1 & c_1 & 0 & 0 & 0 & 0 & \hdots \\
    0 & -a_{\beta} - a_{\alpha} & 0 & 0 & 0 & 0 & \hdots \\
    0 & 0 & -a_{\alpha_1}+ c_2 & c_2 & c_2 & c_2 & \hdots \\
    0 & 0 & 0 & -a_{\alpha_2} - a_{\alpha_1} & -a_{\alpha_1} & -a_{\alpha_1} & \hdots \\
    0 & 0 & 0 & 0 & -a_{\alpha_3} - a_{\alpha_2} & -a_{\alpha_2} & \hdots \\
    \vdots & & & \ddots & & & \hdots
  \end{pmatrix}
\end{equation}

given that $a, c > 0$ for all $a$ and $c$, all pivots are guaranteed to be negative except for,

\begin{align}
    - a_{\alpha} + c_1 \nonumber \\
    -a_{\alpha_1} + c_2
\end{align}

Therefore to obtain a negative definite Hessian, it is sufficient to ascertain that these two terms are negative. That is, the following conditions should be met,

\begin{align}
     \sum_{i=0}^{(\sum_{d=1}^Dx_d)-1} \frac{1}{(\alpha + i)^2} & > \sum_{i=0}^{(\sum_{d=1}^{D+1}x_d)-1} \frac{1}{(\alpha + \beta + i)^2} \nonumber \\
     \sum_{i=0}^{x_1 - 1} \frac{1}{(\alpha_1 + i)^2} & > \sum_{i=0}^{(\sum_{d=1}^Dx_d)-1} \frac{1}{(\sum_{d=1}^D\alpha_d + i)^2}
\end{align}

which are equivalent to,

\begin{align}
     \sum_{i=0}^{(\sum_{d=1}^Dx_d)-1} \left( \frac{1}{(\alpha + i)^2} - \frac{1}{(\alpha + \beta + i)^2}\right)& > \sum_{i=\sum_{d=1}^Dx_d}^{(\sum_{d=1}^{D+1}x_d)-1} \frac{1}{(\alpha + \beta + i)^2} \nonumber \\
     \sum_{i=0}^{x_1 - 1} \left(\frac{1}{(\alpha_1 + i)^2} - \frac{1}{(\sum_{d=1}^D\alpha_d + i)^2} \right) & > \sum_{i=x_1}^{(\sum_{d=1}^Dx_d)-1} \frac{1}{(\sum_{d=1}^D\alpha_d + i)^2}
\end{align}

For the practical application of the BLM in classification, for the purposes of the current paper, we added a small positive number to ensure these conditions are met.

\end{document}


\maketitle

\beginsupplement

\section{Finite Limit Approximations}
We attempted to approximate the following finite sums of the general form,

\begin{equation}\label{eq13}
	f_1(\theta, N) = \sum_{n=0}^N \frac{1}{\theta + n} = \frac{1}{\theta} + \sum_{n=1}^N \frac{1}{\theta + n}
\end{equation}

\begin{equation}\label{eq14}
	f_2(\theta, N) = \sum_{n=0}^N \frac{1}{(\theta + n)^2} = \frac{1}{\theta^2} + \sum_{n=1}^N \frac{1}{(\theta + n)^2}
\end{equation}

for positive real $\theta$ and non-negative integer $N$.  Approximation was performed by starting with the known approximations for the harmonic and geometric series when $\theta \to 0$.  The R "nls" function was then used to model various functional forms by trial and error, continually increasing in accuracy by modeling the residuals of each fit.  The following are the approximations used in the manuscript for equations 13 and 14, respectively,

\begin{equation}
	\hat{f}_1(\theta, N) = \frac{1}{\theta} + \frac{\ln(N) + \gamma + \frac{1}{2N} - \frac{1}{12N^2}}{1 + e^{\left( \frac{\ln(\theta) - \frac{\ln(N)}{2} + \frac{e}{10}}{0.072 \ln(N)^{1.27} + \frac{0.1677}{1 + \ln(N)} + 0.835} \right)}} + A \cos \left( \frac{3 \pi}{2} - \left[ \frac{4 \pi}{1 + e^{ \frac{\ln(\theta) - \frac{\ln(N)}{2} + \frac{e}{10} }{v \left( 0.072 \ln(N)^{1.27} + \frac{0.1677}{1 + \ln(N)} + 0.835 \right)} }} \right] \right)
\end{equation}

where $\gamma$ is the Euler-Mascheroni constant, $A = 0.1068 \left( \ln(N)^{0.8224} + \frac{4.986}{\ln(N) + 6.408} \right) - 0.7751$, and $v = 3.764 e^{ \left( \frac{-\pi}{6} \right) \left( \ln(N) + 1 \right)^{1.06} } + 1.59$,

\begin{equation}
	\hat{f}_2(\theta, N) = \frac{1}{\theta^2} + \frac{\frac{\pi^2}{6}}{1 + \left( \frac{\pi}{2} e^{\theta} \right)}, \forall N > 200
\end{equation}

where $\hat{f}_2$ is only accurate for $N > 200$, and all computations that did not meet this criterion were calculated exactly (not using this approximation).  Readers familiar with the harmonic and geometric series will recognize that $\ln(N) + \gamma + \frac{1}{2N} - \frac{1}{12N^2}$ is the finite series approximation for the harmonic series, and $\frac{\pi^2}{6}$ is the limit of the geometric series (\citealt{sofo2018,merlini2006}).  Examples of the above approximations are implemented in the open source code included with this manuscript (see Data Availability).

\section{Detailed Methods for Markov Chain Monte-Carlo Data Simulation}
Given a vector of counts $\bm{x}_c$ for class $c \in C$ and a vector of fixed, generating parameters $\bm{\theta}_c$,

\begin{align}
	P(\bm{x}_c|\bm{\theta}_c) = \frac{P(\bm{\theta}_c|\bm{x}_c)P(\bm{x}_c)}{P(\bm{\theta}_c)}
\end{align}

Using a Metropolis-Hastings sampling strategy (\citealt{gilks1996}), data were then generated according to the following acceptance probability $P_\alpha$:

\begin{align}
 \alpha & = \text{min} \left\{1,\frac{P(\bm{x}_{i+1}|\bm{\theta}_c)q(\bm{x}_i|\bm{x}_{i+1})}{P(\bm{x}_i|\bm{\theta}_c)q(\bm{x}_{i+1}|\bm{x}_i)}\right\} \nonumber\\
        & = \text{min} \left\{1,\frac{P(\bm{\theta}_c|\bm{x}_{i+1})P(\bm{x}_{i+1})q(\bm{x}_i|\bm{x}_{i+1})}{P(\bm{\theta}_c|\bm{x}_i)P(\bm{x}_i)q(\bm{x}_{i+1}|\bm{x}_i)}\right\}
\end{align}

Taking $q(\bm{x}_{i+1}|\bm{x}_i) = P(\bm{x}_{i+1})$,

\begin{align}
 \alpha & = \text{min} \left\{1,\frac{P(\bm{\theta}_c|\bm{x}_{i+1})}{P(\bm{\theta}_c|\bm{x}_i)}\right\}\nonumber\\
        & = \text{min} \left\{1,\frac{P(\bm{x}_{i+1}|\bm{\theta}_c)/P(\bm{x}_{i+1})}{P(\bm{x}_i|\bm{\theta}_c)/P(\bm{x}_i)}\right\}
\end{align}

We used a multinomial distribution for the proposal $P(\bm{x})$. We then sampled from a distribution $U \sim \text{uniform}(0,1)$ and accept any new sample $\bm{x}_{i+1}$ if $U \leq \alpha$ and we keep $\bm{x}_i$ otherwise. MCMC simulation was performed separately for each class using a Metropolis-Hastings sampling strategy with a multinomial proposal, $P(\bm{x})$, and acceptance-rejection probability,

\begin{equation}\label{accept}
 \alpha = \text{min} \left\{1,\frac{P(\bm{x}_{i+1}|\bm{\theta}_c)/P(\bm{x}_{i+1})}{P(\bm{x}_i|\bm{\theta}_c)/P(\bm{x}_i)}\right\}
\end{equation}

where $P(\bm{x}_{i+1}|\bm{\theta}_c)$ is the likelihood of $\bm{x}_{i+1}$ under the BLM and $P(\bm{x}_{i+1})$ is the likelihood of the data under the proposal.

\section{Support for Classification Methods}

\subsection{Classification Using the Matrix Inner Product}

We construct a test matrix of counts for each feature $\underset{N \times (D+1)}{\bm{T}}$ and a training matrix with columns on the unit simplex $\underset{(D+1) \times C}{\bm{X}}$.

The maximum \textit{a posteriori} (MAP) estimate for a class $c \in C$ is given by:

\begin{equation}
	c_{\text{MAP}} = \underset{c \in C}{\text{argmax}} P(\bm{X}_c | \bm{T}_n)
\end{equation}

Applying Bayes theorem:

\begin{equation}
	c_{\text{MAP}} = \underset{c \in C}{\text{argmax}} \frac{P(\bm{T}_n | \bm{X}_c) P(\bm{X}_c)}{P(\bm{T}_n)}
\end{equation}

Because $P(\bm{T}_n)$ is constant for all $c \in C$, it can be dropped from the denominator resulting in,

\begin{equation}
	c_{\text{MAP}} \propto \underset{c \in C}{\text{argmax}} P(\bm{T}_n | \bm{X}_c) P(\bm{X}_c)
\end{equation}

Considering the count features (counts of words) in the test vector $\bm{T}_n$ we have,

\begin{equation}
	c_{\text{MAP}} \propto \underset{c \in C}{\text{argmax}} P(t_{n1}, \dots, t_{n(D+1)} | \bm{X}_c) P(\bm{X}_c)
\end{equation}

we make the naive Bayes assumption that all features are conditionally independent of one another results in,

\begin{equation}
	P(t_{n1}, \dots, t_{n(D+1)} | \bm{X}_c) = P(t_{n1}|\bm{X}_c) P(t_{n2}|\bm{X}_c) \dots P(t_{n(D+1)}|\bm{X}_c)
\end{equation}

Assuming that all classes have equal probability further simplifies the calculation of the naive Bayes (NB) estimate:

\begin{equation}
	c_{\text{NB}} \propto \underset{c \in C}{\text{argmax}} \prod_{d=1}^{D+1} P(t_{nd} | \bm{X}_c)
\end{equation}

For a single test count (where $t_{nd} = 1$), $P(t_{nd} | \bm{X}_c) = x_{dc}$, and assuming all test counts are independent draws:

\begin{equation}
	c_{\text{NB}} \propto \underset{c \in C}{\text{argmax}} \prod_{d=1}^{D+1} x_{dc}^{t_{nd}}
\end{equation}

Taking the log of the training matrix $\bm{X}$,

\begin{equation}
	c_{\text{NB}} \propto \underset{c \in C}{\text{argmax}} \sum_{d=1}^{D+1} t_{nd} \ln(x_{dc})
\end{equation}

Which is the definition of the matrix inner product:

\begin{equation}
	\bm{c}_{\text{NB}} \propto \underset{c \in C}{\text{argmax}} \text{ } \bm{T} \ln(\bm{X})
\end{equation}

\subsection{Proof for the Use of BLM MLEs for Classification on the Multinomial Simplex}
Let there be a $D+1$ dimensional vector containing a single count at element $d$, $\bm{y} = \{0,0,0,\dots,1,\dots,0\}$.  We aim to assign this vector to a class $c \in C$. Assume that we have the data matrix $\bm{X} = \{\bm{x}_1, \bm{x}_2, \dots, \bm{x}_C\}$. Such that $\bm{x}_c = \{x_{c1}, \dots,x_{cd},\dots, x_{C(D+1)}\}$. Let $\bm{\theta}$ be the $D+2$ paramters of a BLM distribution and $\bm{p}$ be the $D+1$ paramters of a multinomial distribution. Initially, we assume $\bm{\theta}$ are known.  Additionally, we assume a $0/1$ loss function that will assign $\bm{y}$ to class $c$ if $P(c|\bm{y},\bm{X})$ is maximum. Accordingly, we want to find,

\begin{equation}\label{eq26}
 P(c|\bm{y},\bm{X},\bm{\theta}) = \frac{P(\bm{y}|c,\bm{X},\bm{\theta})P(c|\bm{X},\bm{\theta})}{\sum_C P(\bm{y}|c,\bm{X},\bm{\theta})P(c|\bm{X},\bm{\theta})}
\end{equation}

Assuming that the classes have equal priors that do not depend on the data and that the classes are independent,

\begin{equation}\label{eq27}
 P(c|\bm{y},\bm{X},\bm{\theta}) = \frac{P(\bm{y}|\bm{x}_c,\bm{\theta}_c)}{\sum_{C}P(\bm{y}|\bm{x}_c,\bm{\theta}_c)}
\end{equation}

Given that the denominator of \ref{eq27} is constant,

\begin{equation}\label{eq28}
 P(c|\bm{y},\bm{X},\bm{\theta}) \propto P(\bm{y}|\bm{x}_c,\bm{\theta}_c) 
\end{equation}

Assuming that we have produced BLM MLEs, $\hat{\bm{\theta}_c}$, utilizing the data $\bm{x}_c$, and that they are sufficient statistics then,

\begin{equation}\label{eq29}
 P(c|\bm{y},\bm{X},\bm{\theta}) \propto P(\bm{y}|\hat{\bm{\theta}_c}) 
\end{equation}

Using equation 4 in the manuscript Section 2.2 we find,

\begin{equation}\label{eq30}
   P(\bm{y}|\hat{\bm{\theta}_c}) = \frac{\Gamma\left(\sum_{d=1}^D\hat{\alpha}_{cd}  
                                   \right)\Gamma(\hat{\alpha}_c+\hat{\beta}_c)}{\Gamma(\hat{\alpha}_c)\Gamma(\hat{\beta}_c)\prod_{d=1}^D\Gamma(\hat{\alpha}_{cd})}  \frac{\Gamma\left(\left(\sum_{d=1}^{D+1}y_d\right)+1\right)}{\prod_{d=1}^{D+1}\Gamma(y_d+1)}\frac{\Gamma(\hat{\alpha}_c+\sum_{d=1}^D y_d)\Gamma(\hat{\beta}_c+y_{D+1})}{\Gamma(\hat{\alpha}_c+\hat{\beta}_c+\sum_{d=1}^{D+1}y_d)}\frac{\prod_{d=1}^D\Gamma(\hat{\alpha}_{cd}+y_d)}{\Gamma(\sum_{d=1}^D\hat{\alpha}_{cd}+y_d)}
\end{equation}

For the $\bm{y} = \{0,0,0, \dots , 1, \dots,0,0,0\}$ with $y_d = 1$ and by noting that $\Gamma(n+1)=n\Gamma(n)$ we get, for $d \neq D+1$,

\begin{equation}\label{eq31}
   P(y_d|\hat{\bm{\theta}_c}) = \hat{p}_d = \frac{\hat{\alpha}_c}{\hat{\alpha}_c+\hat{\beta}_c}\frac{\hat{\alpha}_{cd}}{\sum_{d=1}^D\hat{\alpha}_{cd}}
\end{equation}

and for $d = D+1$,

\begin{equation}\label{eq32}
   P(y_d|\hat{\bm{\theta}_c}) = \hat{p}_{D+1} = \frac{\hat{\beta}_c}{\hat{\alpha}_c+\hat{\beta}_c}
\end{equation}

We can now use these as the parameters $\bm{p}$ in a multinomial, assuming independence of the vectors $\bm{y}$, which results in, 

\begin{equation}\label{eq33}
 P(\bm{y}|\hat{\bm{\theta}_c}) = \left(\frac{\hat{\beta}_c}{\hat{\alpha}_c+\hat{\beta}_c}\right)^{y_{D+1}}\prod_{d=1}^D\left(\frac{\hat{\alpha}_c}{\hat{\alpha}_c+\hat{\beta}_c}\frac{\hat{\alpha}_{cd}}{\sum_{d=1}^D\hat{\alpha}_{cd}}\right)^{y_d}
\end{equation}

where $y_d \geq 0$.

\section{Gold Standard Data Preprocessing}

All datasets were preprocessed by Cardoso-Cachopo (\citeyear{cardoso2007-1}) as follows:

\begin{itemize}
	\item Tab, newline, and return characters were replaced with a space.  
	\item Only letters were kept; special characters and numbers were removed.
	\item All letters were modified to lowercase.
\end{itemize}

Additionally, all English datasets (excluding Cade12) were preprocessed as follows:

\begin{itemize}
	\item Words less than 3 characters in length were removed.
	\item All 524 English stopwords from the SMART Retrieval System were removed (Salton 1971).
	\item The remaining words were processed using the Porter Stemming Algorithm (Rijsbergen et al. 1980).
\end{itemize}

The post-processed datasets are available in the open source GitHub repository associated with this publication (see Data Availability).

\section{Supplementary Figures}

\begin{figure}[h!]
  \centering
  \includegraphics[width=\textwidth]{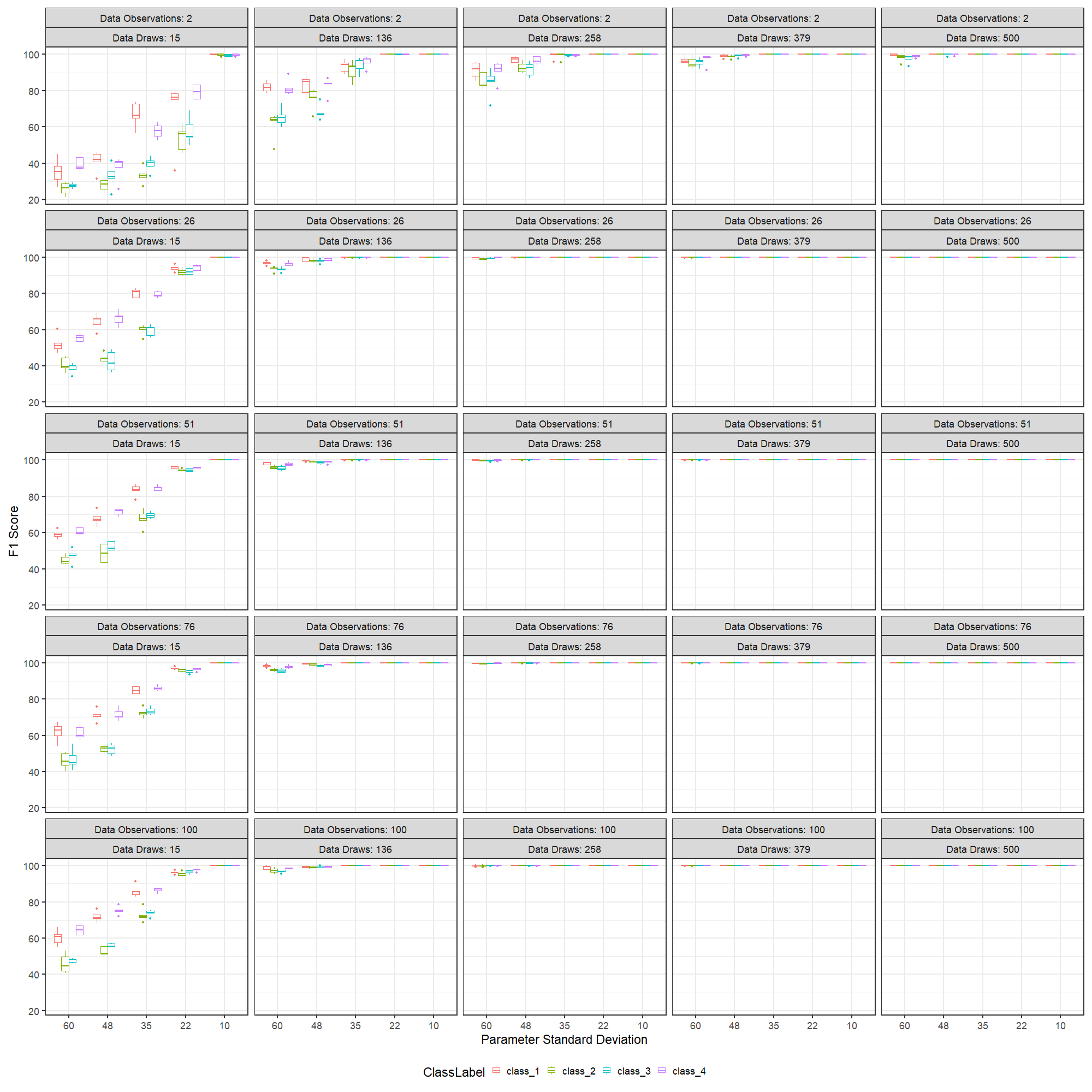}
  \caption{\textbf{Use of the DM distribution results in high F1 classification scores as data draws and observations increase, even for difficult classification tasks.}  A separate boxplot is shown for each simulated class distribution (classes 1-4).  The boxplots summarize the F1 score quartiles for 5 independent datasets for each combination of variables.}
  \label{fig:sfig1}
\end{figure}

\begin{figure}[h!]
  \centering
  \includegraphics[width=\textwidth]{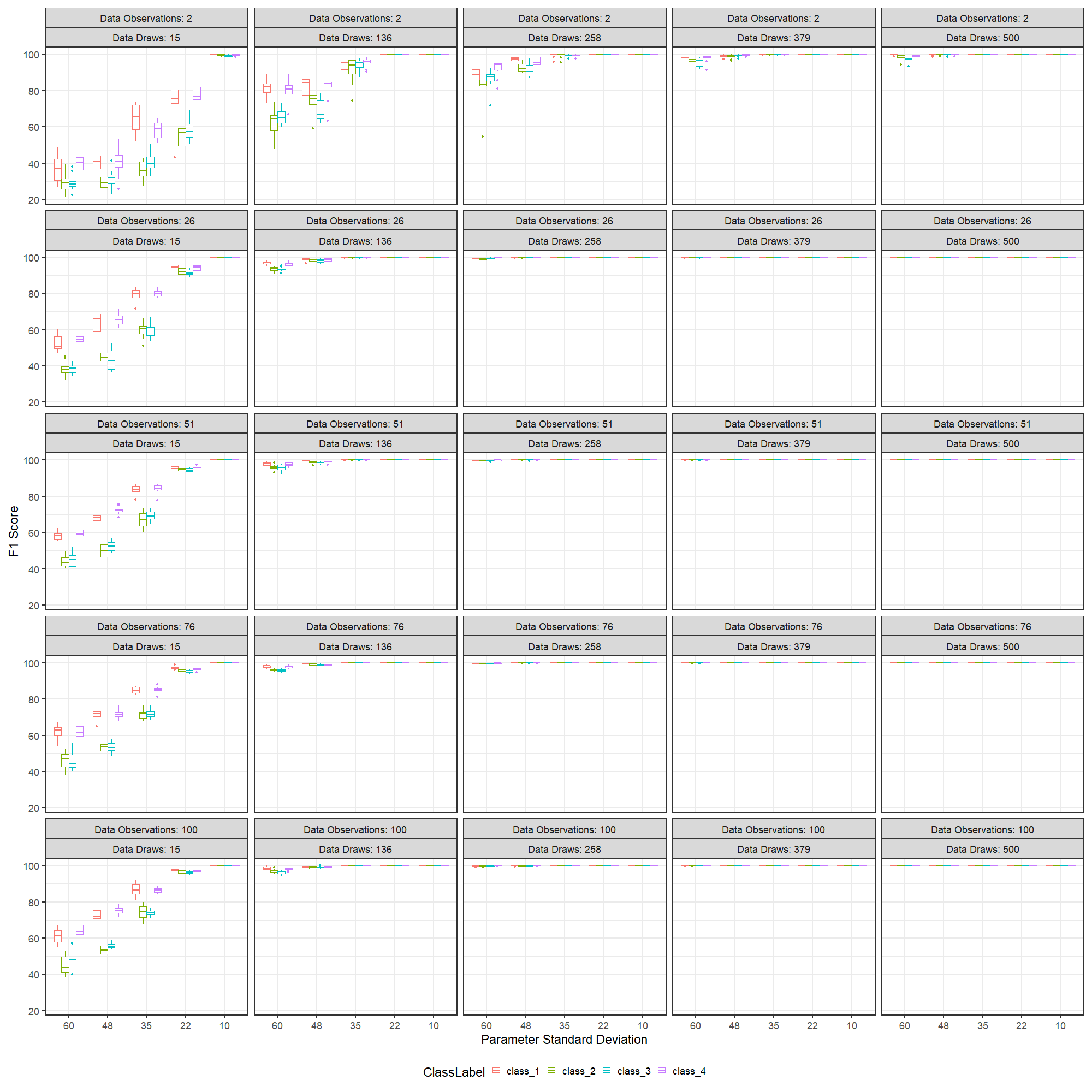}
  \caption{\textbf{Use of the multinomial distribution and standard multinomial naive Bayes results in high F1 classification scores as data draws and observations increase, even for difficult classification tasks.}  A separate boxplot is shown for each simulated class distribution (classes 1-4).  The boxplots summarize the F1 score quartiles for 5 independent datasets for each combination of variables.}
  \label{fig:sfig2}
\end{figure}

\begin{figure}[h!]
  \centering
  \includegraphics[width=\textwidth]{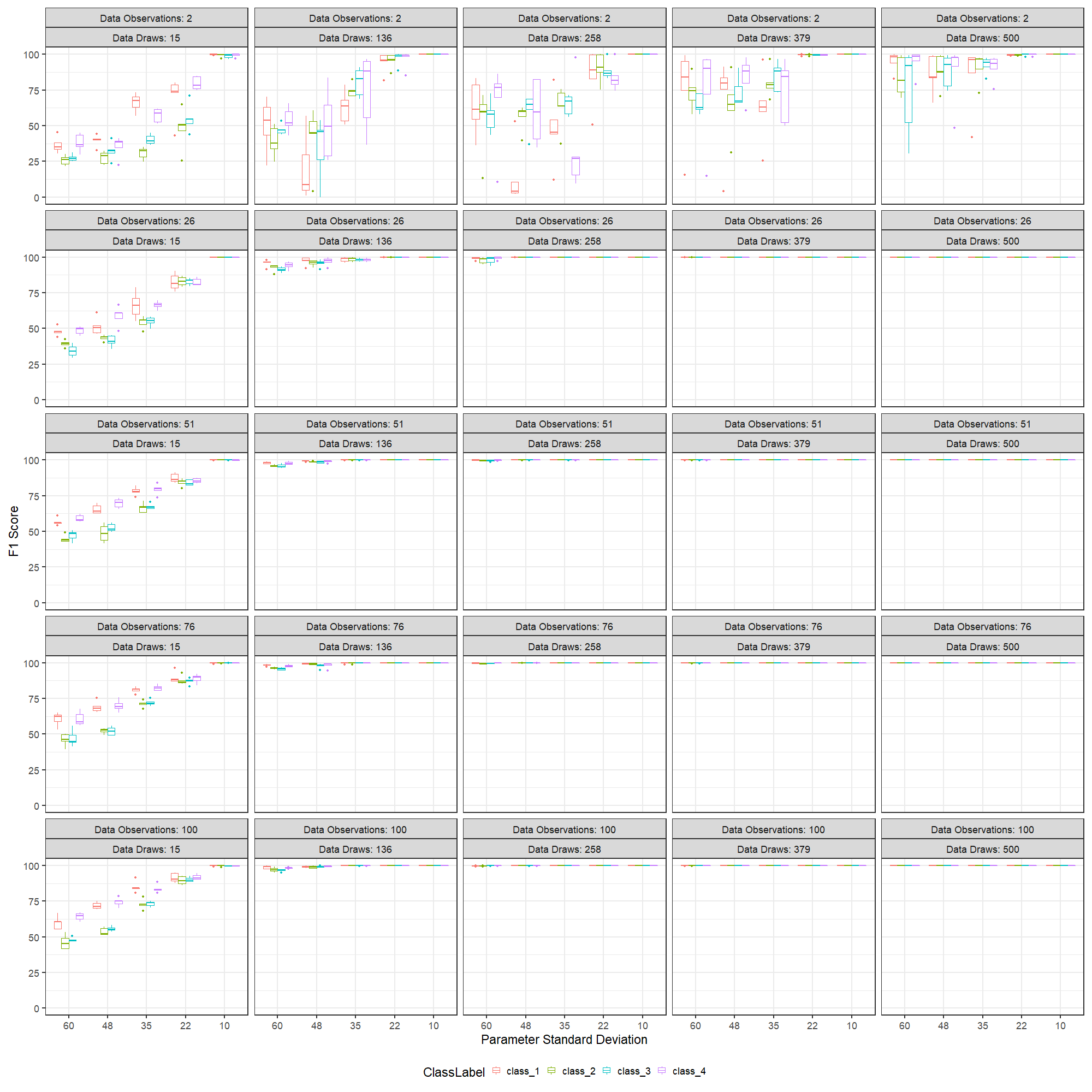}
  \caption{\textbf{Use of the DM distribution with marginal likelihood classification results in high F1 classification scores as data draws and observations increase, even for difficult classification tasks.}  However, with the marginal likelihood classification methods, more data is required to produce classification results with similar F1 scores to the naive Bayes approaches.  A separate boxplot is shown for each simulated class distribution (classes 1-4).  The boxplots summarize the F1 score quartiles for 5 independent datasets for each combination of variables.}
  \label{fig:sfig3}
\end{figure}

\begin{figure}[h!]
  \centering
  \includegraphics[width=\textwidth]{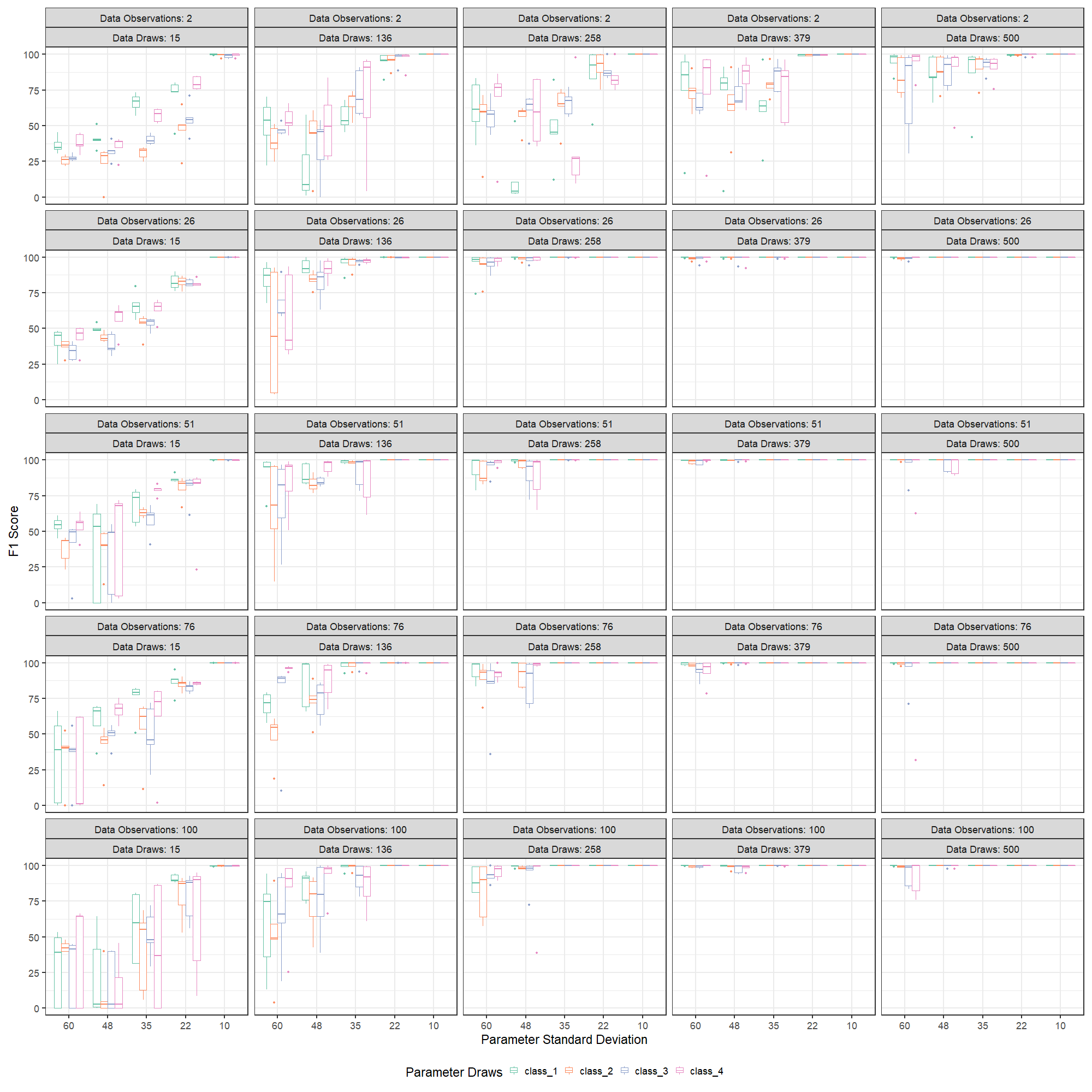}
  \caption{\textbf{Use of the BLM distribution with marginal likelihood classification results in high F1 classification scores as data draws and observations increase, even for difficult classification tasks.}  However, with the marginal likelihood classification methods, more data is required to produce classification results with similar F1 scores to the naive Bayes approaches.  A separate boxplot is shown for each simulated class distribution (classes 1-4).  The boxplots summarize the F1 score quartiles for 5 independent datasets for each combination of variables.}
  \label{fig:sfig4}
\end{figure}

\begin{figure}[h!]
  \centering
  \includegraphics[width=\textwidth]{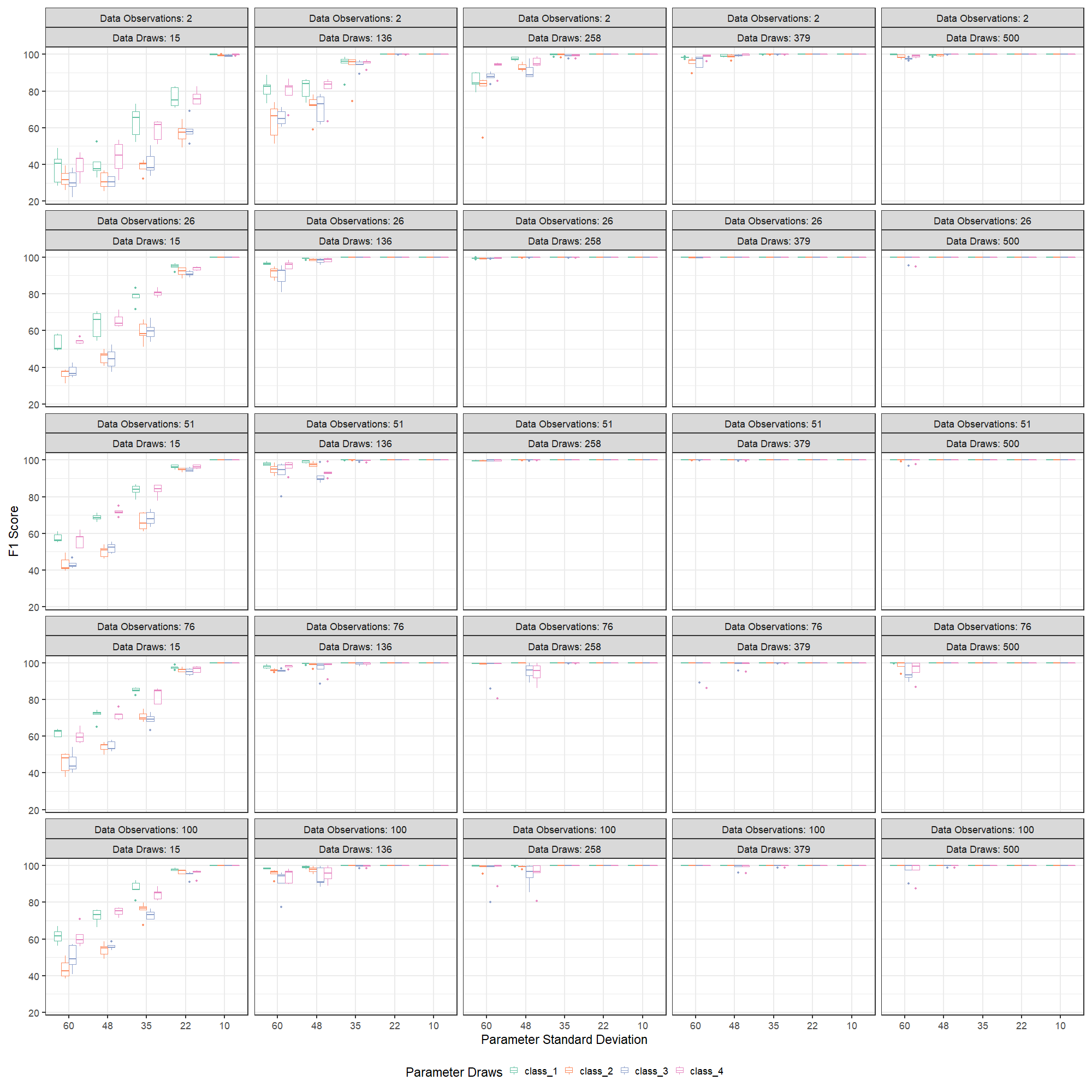}
  \caption{\textbf{On data simulated from the BLM distribution, use of the BLM distribution results in high F1 classification scores as data draws and observations increase, even for difficult classification tasks.}  There doesn't appear to be a difference between classification scores on multinomial data compared to BLM-simulated data.  A separate boxplot is shown for each simulated class distribution (classes 1-4).  The boxplots summarize the F1 score quartiles for 5 independent datasets for each combination of variables.}
  \label{fig:sfig5}
\end{figure}

\begin{figure}[h!]
  \centering
  \includegraphics[width=\textwidth]{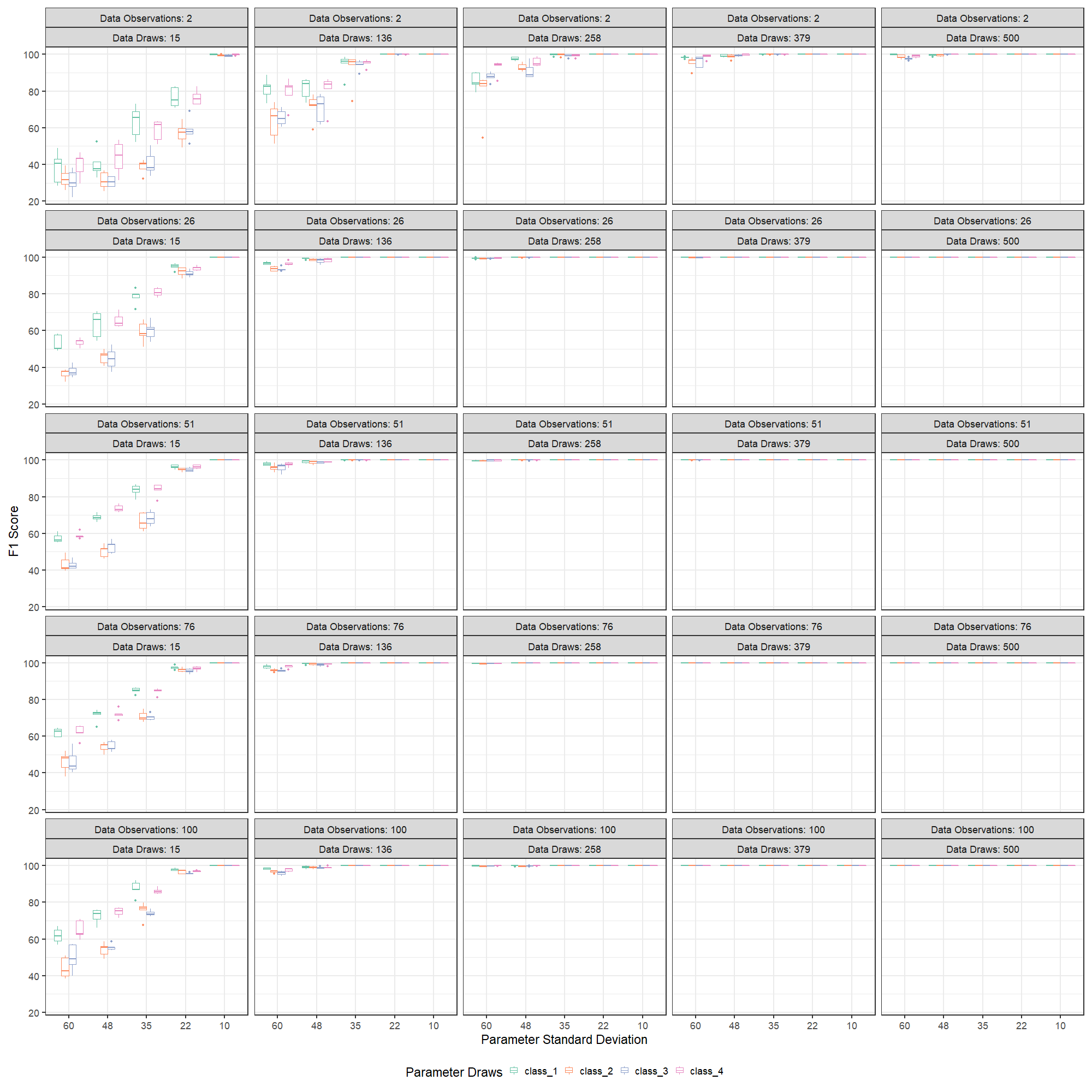}
  \caption{\textbf{On data simulated from the BLM distribution, use of the DM distribution results in high F1 classification scores as data draws and observations increase, even for difficult classification tasks.}  There doesn't appear to be a difference between classification scores on multinomial data compared to BLM-simulated data.  A separate boxplot is shown for each simulated class distribution (classes 1-4).  The boxplots summarize the F1 score quartiles for 5 independent datasets for each combination of variables.}
  \label{fig:sfig6}
\end{figure}

\clearpage
\bibliographystyle{apalike}
\bibliography{LakinAbdo_FastMLEBetaLiouville_SupplementaryFile1}